\documentclass[letterpaper]{nature}
\usepackage{color}
\usepackage{xurl}
\usepackage{verbatim}
\usepackage{multirow}
\usepackage{xspace}
\usepackage{graphicx}
\usepackage{amsmath} 
\usepackage{bbm}
\usepackage{amssymb}
\usepackage{booktabs} 
\usepackage{times} 
\usepackage{lscape}
\usepackage[left=1in,right=1in,bottom=1.1in,top=1in]{geometry}
\usepackage{enumitem}
\usepackage[table,xcdraw]{xcolor}
\usepackage[normalem]{ulem}
\usepackage{xcolor}
\usepackage[innercaption]{sidecap}
\usepackage{lineno}
\usepackage{enumitem}
\usepackage{xcolor}

\usepackage{xr-hyper}
\makeatletter
\newcommand*{\addFileDependency}[1]{% argument=file name and extension
  \typeout{(#1)}
  \@addtofilelist{#1}
  \IfFileExists{#1}{}{\typeout{No file #1.}}
}
\makeatother

\usepackage[colorlinks,citecolor=blue,urlcolor=gray]{hyperref}
\usepackage[margin=-1cm, labelfont=bf, font=footnotesize]{caption}
\DeclareCaptionLabelFormat{adja-page}{\hrulefill\\#1 #2 \emph{(previous page)}}
\usepackage{setspace}

\usepackage{algpseudocode}
\usepackage[american]{babel}
\usepackage{microtype}
\usepackage{graphicx}
\usepackage{subcaption}
\usepackage{booktabs}
\usepackage{multirow}
\usepackage{paralist}
\usepackage{bbm}
\usepackage{hyperref}
\usepackage{amssymb}
\usepackage{mathtools}
\usepackage[ruled,vlined]{algorithm2e}
\usepackage{algpseudocode}
\usepackage{mathtools}
\usepackage{booktabs}
\usepackage{tikz}
\usepackage{enumitem}
\usepackage{makecell}

\mathchardef\mhyphen="2D

\newcommand{\vertiii}[1]{{\left\vert\kern-0.25ex\left\vert\kern-0.25ex\left\vert #1
    \right\vert\kern-0.25ex\right\vert\kern-0.25ex\right\vert}}

\def\bM{{\mathbf{M}}}

\def\bX{{\mathbf{X}}}
\def\bY{{\mathbf{Y}}}

\def\cG{\mathcal{G}}

\def\cK{\mathcal{K}}

\def\cS{\mathcal{S}}

\DeclareMathOperator*{\argmin}{arg\,min}

\usepackage{pifont}
\definecolor{lightgreen}{rgb}{0.47, 1.0, 0.47}
\definecolor{darkgreen}{rgb}{0., 0.6, 0.}
\definecolor{lightpurple}{rgb}{.84, 0.55, 0.97}
\definecolor{darkpurple}{rgb}{.51, 0.04, 0.73}
\definecolor{lightred}{rgb}{1., 0.6, 0.68}
\definecolor{darkred}{rgb}{1.0, 0., 0.2}
\definecolor{bluefeatlow}{HTML}{F89658}
\definecolor{bluefeatmid}{HTML}{EF7021}
\definecolor{bluefeathigh}{HTML}{AF4503}
\newcommand{\std}[1]{\scriptsize{$\pm$#1}}

\newcommand{\datagen}{\textsc{ShapeGGen}\xspace}
\newcommand{\name}{\textsc{GraphXAI}\xspace}
\newcommand{\sgbase}{\textsc{SG-Base}\xspace}
\newcommand{\sghetero}{\textsc{SG-Heterophilic}\xspace}
\newcommand{\sgunfair}{\textsc{SG-Unfair}\xspace}
\newcommand{\sgsmallex}{\textsc{SG-SmallEx}\xspace}
\newcommand{\sgmoreinform}{\textsc{SG-MoreInform}\xspace}
\newcommand{\sglessinform}{\textsc{SG-LessInform}\xspace}

\newcommand{\hide}[1]{}
\newcommand{\xhdr}[1]{\vspace{1.7mm}\noindent{{\bf #1.}}}

\newcommand{\eg}{\emph{e.g.}\xspace}
\newcommand{\ie}{\emph{i.e.}\xspace}

\let\oldnl\nl
\newcommand{\nonl}{\renewcommand{\nl}{\let\nl\oldnl}}  
\usepackage[normalem]{ulem}
\usepackage{comment}
\usepackage{bm}

\title{\begin{center}
Evaluating explainability for graph neural networks
\end{center}}    

\author  
{\begin{center}   
Chirag~Agarwal$^{1,4, \star}$, Owen~Queen$^{4,2, \star}$, Himabindu Lakkaraju$^{3,6,7}$, and Marinka Zitnik$^{4,5,6,\ddag}$ \\[1mm]  
\normalsize{$^{1}$Media and Data Science Research Lab, Adobe, Noida 201304, India}\\
\normalsize{$^{2}$Department of Computer Science, University of Tennessee, Knoxville, TN 37996, USA} \\
\normalsize{$^{3}$Harvard Business School, Boston, MA 02163, USA} \\
\normalsize{$^{4}$Department of Biomedical Informatics, Harvard University, Boston, MA 02115, USA} \\  
\normalsize{$^{5}$Broad Institute of MIT and Harvard, Cambridge, MA 02142, USA} \\
\normalsize{$^{6}$Harvard Data Science Initiative, Cambridge, MA 02138, USA} \\
\normalsize{$^{7}$Department of Computer Science, Harvard University, Boston, MA 02134, USA}\\ [4mm]
\normalsize{$^\star$ Equal contribution} \\ 
\normalsize{$\ddag$ Corresponding author: marinka@hms.harvard.edu}
\end{center}
}

\begin{document}

\maketitle

\singlespacing

% {
\spacing{1.4} %1.65
\begin{abstract}
\noindent As explanations are increasingly used to understand the behavior of graph neural networks (GNNs), evaluating the quality and reliability of GNN explanations is crucial. However, assessing the quality of GNN explanations is challenging as existing graph datasets have no or unreliable ground-truth explanations. Here, we introduce a synthetic graph data generator, \datagen, which can generate a variety of benchmark datasets (e.g., varying graph sizes, degree distributions, homophilic vs. heterophilic graphs) accompanied by ground-truth explanations. The flexibility to generate diverse synthetic datasets and corresponding ground-truth explanations allows \datagen to mimic the data in various real-world areas. We include \datagen and several real-world graph datasets in a graph explainability library, \name. In addition to synthetic and real-world graph datasets with ground-truth explanations, \name provides data loaders, data processing functions, visualizers, GNN model implementations, and evaluation metrics to benchmark GNN explainability methods.
\end{abstract}
% }

\clearpage

\spacing{1.38}

% !TEX root = main.tex
\section*{Introduction}
As graph neural networks (GNNs) are being increasingly used for learning representations of graph-structured data in high-stakes applications, such as criminal justice\cite{agarwal2021towards}, molecular chemistry\cite{sanchez2020evaluating,giunchiglia2022towards}, and biological networks\cite{morselli2021network,zitnik2018modeling}, it becomes critical to ensure that the relevant stakeholders can understand and trust their functionality. To this end, previous work developed several methods to explain predictions made by GNNs\cite{baldassarre2019explainability,faber2020contrastive,huang2022graphlime,lucic2022cf,luo2020parameterized,pope2019explainability,schlichtkrull2020interpreting,vu2020pgm,ying2019gnnexplainer}. With the increase in newly proposed GNN explanation methods, it is critical to ensure their reliability. However, explainability in graph machine learning is an emerging area lacking standardized evaluation strategies and reliable data resources to evaluate, test, and compare GNN explanations\cite{agarwal2021probing}. While several works have acknowledged this difficulty, they tend to base their analysis on specific real-world\cite{sanchez2020evaluating} and synthetic\cite{faber2021comparing} datasets with limited ground-truth explanations. In addition, relying on these datasets and associated ground-truth explanations is insufficient as they are not indicative of diverse real-world applications\cite{agarwal2021probing}. To this end, developing a broader ecosystem of data resources for benchmarking state-of-the-art GNN explainers can support explainability research in GNNs.

A comprehensive data resource to correctly evaluate the quality of GNN explanations will ensure their reliability in high-stake applications. However, the evaluation of GNN explanations is a growing research area with relatively little work, where existing approaches mainly leverage ground-truth explanations associated with specific datasets\cite{sanchez2020evaluating} and are prone to several pitfalls (as outlined by Faber et al.\cite{faber2021comparing}). Further, multiple underlying rationales can generate the correct class labels, creating redundant or non-unique explanations. A trained GNN model may only capture one or an entirely different rationale. In such cases, evaluating the explanation output by a state-of-the-art method using the ground-truth explanation is incorrect because the underlying GNN model does not rely on that ground-truth explanation. In addition, even if a unique ground-truth explanation generates the correct class label, the GNN model trained on the data could be a weak predictor using an entirely different rationale for prediction. Therefore, the ground-truth explanation cannot be used to assess post hoc explanations of such models. Lastly, the ground-truth explanations corresponding to some of the existing benchmark datasets are not good candidates for reliably evaluating explanations as they can be recovered using trivial baselines (\eg, random node or edge as explanation). The above discussion highlights a clear need for general-purpose data resources which can evaluate post hoc explanations reliably across diverse real-world applications. While various benchmark datasets (\eg, Open Graph Benchmark (OGB)\cite{hu2020ogb}, GNNMark\cite{baruah2021gnnmark}, GraphGT\cite{du2021graphgt}, MalNet\cite{freitas2021large}, Graph Robustness Benchmark (GRB)\cite{zheng2021graph}, Therapeutics Data Commons\cite{huang2021therapeutics,huang2022artificial}, and EFO-1-QA\cite{wang2021benchmarking}) and programming libraries for deep learning on graphs (\eg, Dive Into Graphs (DIG)\cite{dig2021}, Pytorch Geometric (PyG)\cite{fey2019fast}, and Deep Graph Library (DGL)\cite{wang2019deep}) in graph machine learning literature exist, they are mainly used to only benchmark the performance of GNN predictors and are not suited to evaluate the correctness of GNN explainers because they do not capture ground-truth explanations.

\begin{figure*}[ht]
    \centering
    \includegraphics[width=0.99\textwidth]{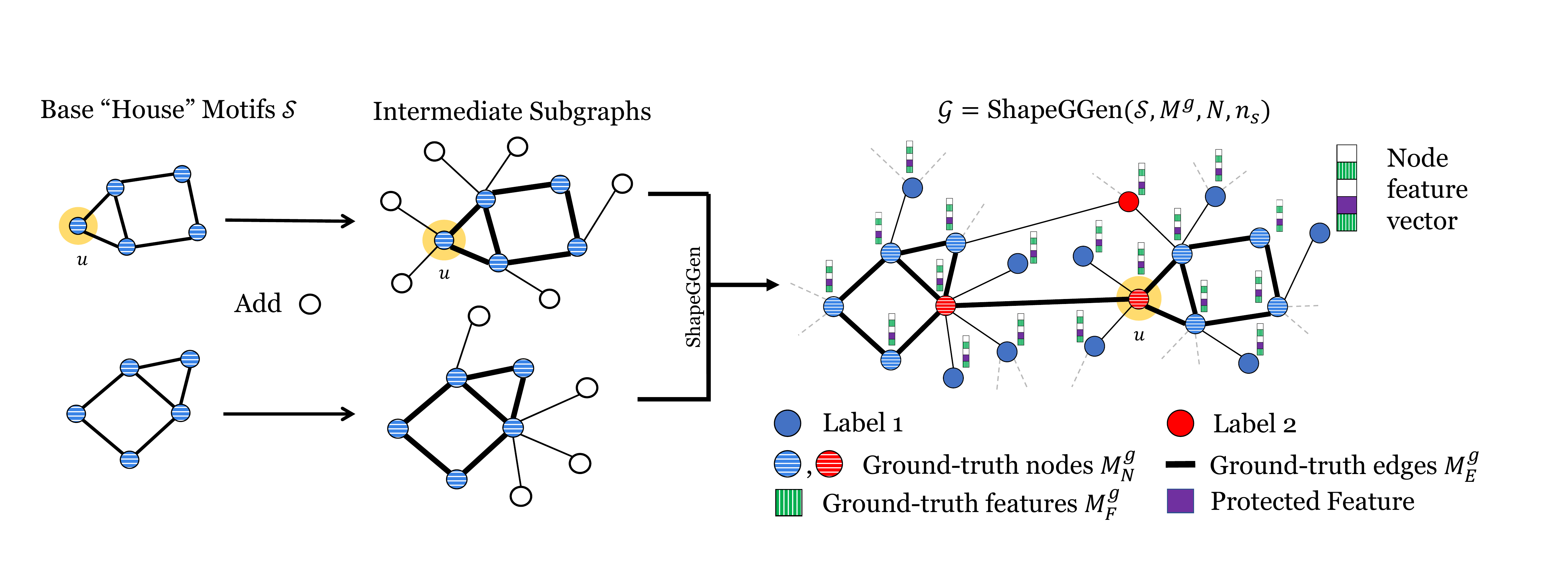}
    \caption{\textbf{Overview of \datagen graph dataset generator.} 
    \datagen is a novel dataset generator for graph-structured data that can be used to benchmark graph explainability methods using ground-truth explanations. Graphs are created by combining subgraphs containing any given motif and additional nodes. The number of motifs in a $k$-hop neighborhood determines the node label (in the figure, we use a 1-hop neighborhood for labeling, and nodes with two motifs in its 1-hop neighborhood are highlighted in red). Feature explanations are some masks over important node features (green striped), with an option to add a protected feature (shown in purple) whose correlation to node labels is controllable. Node explanations are nodes contained in the motifs (horizontal striped nodes) and edge explanations (bold lines) are edges connecting nodes within motifs. 
    }
    \label{fig:shapegraph}
\end{figure*}

Here, we address the above challenges by introducing a general-purpose data resource that is not prone to ground-truth pitfalls (\eg, redundant explanations, weak GNN predictors, trivial explanations) and can cater to diverse real-world applications. To this end, we present \datagen (Figure~\ref{fig:shapegraph}), a novel and flexible synthetic XAI-ready (explainable artificial intelligence ready) dataset generator which can automatically generate a variety of benchmark datasets (\eg, varying graph sizes, degree distributions, homophilic vs. heterophilic graphs) accompanied by ground-truth explanations. \datagen ensures that the generated ground-truth explanations are not prone to the pitfalls described in Faber et al.\cite{faber2021comparing}, such as redundant explanations, weak GNN predictors, and trivially correct explanations. Furthermore, \datagen can evaluate the goodness of any explanation (\eg, node feature-based, node-based, edge-based) across diverse real-world applications by seamlessly generating synthetic datasets that can mimic the properties of real-world data in various domains. 

\begin{figure*}[ht]
    \centering
    \includegraphics[width=0.99\textwidth]{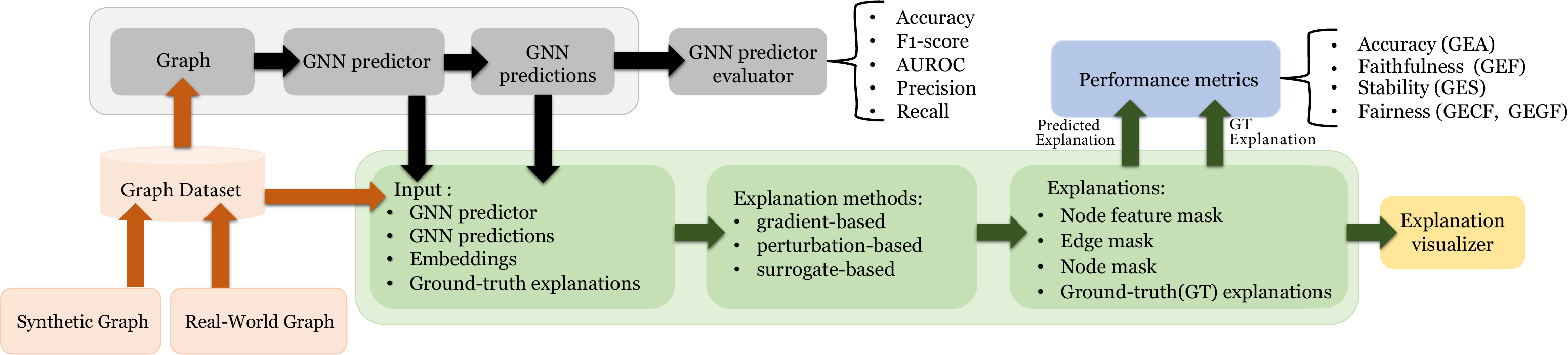}
    \caption{\textbf{Overview of \name.} \name provides data loader classes for XAI-ready synthetic and real-world graph datasets with ground-truth explanations for evaluating GNN explainers, implementations of explanation methods, visualization functions for GNN explainers, utility functions to support new GNN explainers, and a diverse set of performance metrics to evaluate the reliability of explanations generated by GNN explainers.
    }
    \label{fig:graphxai-pipeline}
\end{figure*}

We incorporate \datagen and several other synthetic and real-world graphs~\cite{graphXAI22} into \name, a general-purpose framework for benchmarking GNN explainers. \name also provides a broader ecosystem (Figure~\ref{fig:graphxai-pipeline}) of data loaders, data processing functions, visualizers, and a set of evaluation metrics (\eg, accuracy, faithfulness, stability, fairness) to reliably benchmark the quality of any given GNN explanation (node feature-based or node/edge-based). We leverage various synthetic and real-world datasets and evaluation metrics from \name to empirically assess the quality of explanations output by eight state-of-the-art GNN explanation methods. Across many GNN explainers, graphs, and graph tasks, we observe that state-of-the-art GNN explainers fail on graphs with large ground-truth explanations (\ie, explanation subgraphs with a higher number of nodes and edges) and cannot produce explanations that preserve fairness properties of underlying GNN predictors.

\clearpage

\section*{Results}
To evaluate \name, we show how \name enables systematic benchmarking of eight state-of-the-art GNN explainers on both \datagen (in the Methods section) and real-world graph datasets. We explore the utility of the \datagen generator to benchmark GNN explainers on graphs with homophilic vs. heterophilic, small vs. large,  and fair vs. unfair ground-truth explanations. Additionally, we examine the utility of GNN explanations on datasets with varying degrees of informative node features. Next, we outline the experimental setup, including details about performance metrics, GNN explainers, and underlying GNN predictors, and proceed with a discussion of benchmarking results.

\subsection*{Experimental setup}
\xhdr{GNN explainers}
The \name defines an Explanation class capable of storing multiple types of explanations produced by GNN explainers and provides a \texttt{graphxai.BaseExplainer} class that serves as the base for all explanation methods in \name. We incorporate eight GNN explainability methods, including gradient-based: Grad\cite{simonyan2013saliency}, GradCAM\cite{pope2019explainability}, GuidedBP\cite{baldassarre2019explainability}, Integrated Gradients\cite{sundararajan2017axiomatic}; perturbation-based: GNNExplainer\cite{ying2019gnnexplainer}, PGExplainer\cite{luo2020parameterized}, SubgraphX\cite{yuan2021explainability}; and
surrogate-based methods: PGMExplainer\cite{vu2020pgm}. Finally, following Agarwal et al.\cite{agarwal2021probing}, we consider random explanations as a reference: 1) Random Node Features, a node feature mask defined by an $d$-dimensional Gaussian distributed vector; 2) Random Nodes, a $1\times n$ node mask is randomly sampled using a uniform distribution, where $n$ is the number of nodes in the enclosing subgraph; and 3) Random Edges, an $N\times N$ edge mask drawn from a uniform distribution over a node's incident edges.

\xhdr{Implementation details}
We use a three-layer GIN model\cite{xu2018powerful} and a GCN model \cite{kipf17:semi} as GNN predictors for our experiments. We use a model comprising three GIN convolution layers with ReLU non-linear activation function and a Softmax activation for the final layer. The hidden dimensionality of the layers is set to 16. We follow an established approach for generating explanations\cite{huang2022graphlime,agarwal2021probing} and use reference algorithm implementations. We select top-$k$ ($k=25\%$) important nodes, node features, or edges and use them to generate explanations for all graph explainability methods. For training GIN models, we use an Adam optimizer with a learning rate of $1\times10^{-2}$, weight decay of $1\times10^{-5}$, and the number of epochs to 1000. We use an Adam optimizer with a learning rate of $3\times10^{-2}$, no weight decay, and 1500 training epochs for training GNN models. We set hyperparameters of GNN explainability models following the authors’ recommendations.

We use a fixed random split provided within the \name package to split the datasets. For each \datagen dataset, we use a 70/5/25 split for training, validation, and testing, respectively. For MUTAG, Benzene, and Fluoride Carbonyl datasets, we use a 70/10/20 split throughout each dataset. Average performance is reported across each sample in the testing set of each dataset.

\subsection*{Performance metrics}

\begin{figure}[ht]
    \centering
    \includegraphics[width=0.49\textwidth]{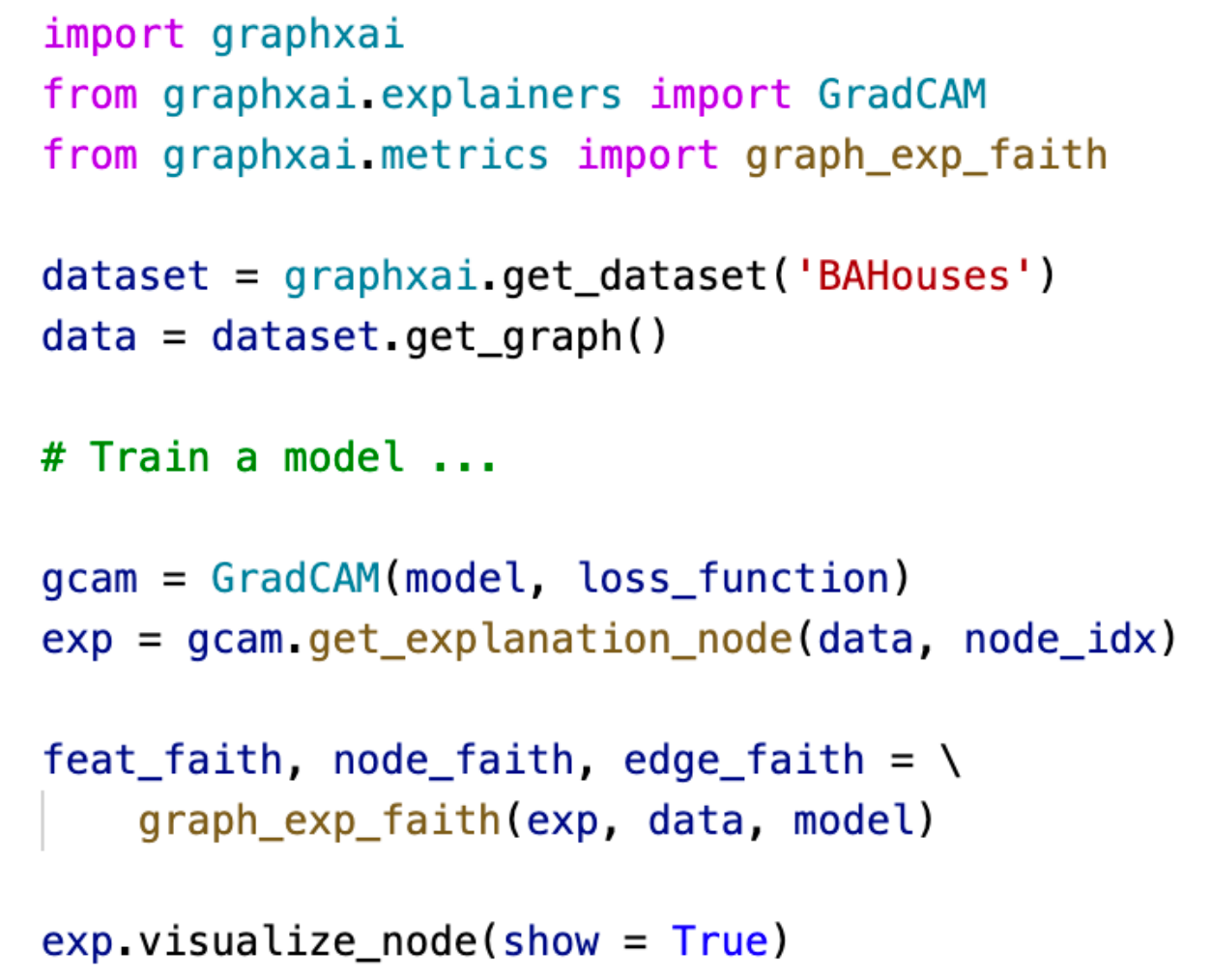}
    \caption{\textbf{Example use case of the \name package.} An example of explaining a prediction in the \name package. With just a few lines of code, one can calculate an explanation for a node or graph, calculate metrics based on that explanation, and visualize the explanation.}
    \label{fig:my_label}
\end{figure}

In addition to the synthetic and real-world data resources, we consider four broad categories of performance metrics: i) Graph Explanation Accuracy (GEA); ii) Graph Explanation Faithfulness (GEF); iii) Graph Explanation Stability (GES); and iv) Graph Explanation Fairness (GECF, GEGF) to evaluate the explanations on the respective datasets. In particular, all evaluation metrics leverage predicted explanations, ground-truth explanations, and other user-controlled parameters, such as top-$k$ features. Our \name package implements these performance metrics and additional utility functions within \texttt{graphxai.metrics} module. Figure~\ref{fig:my_label} shows a code snippet for evaluating the correctness of output explanations for a given GNN prediction in \name.

\xhdr{Graph explanation accuracy (GEA)} We report graph explanation accuracy as an evaluation strategy that measures an explanation's correctness using the ground-truth explanation $\bM^g$. In ground-truth and predicted explanation masks, every node, node feature, or edge belongs to $\{0, 1\}$, where `0' means that an attribute is unimportant and `1' means that it is important for the model prediction. To measure accuracy, we use Jaccard index\cite{taha2015metrics} between the ground-truth $\bM^g$ and predicted $\bM^p$:
\begin{align}
    \textrm{JAC}(\mathbf{M}^g, \mathbf{M}^p) = \frac{\textrm{TP}(\mathbf{M}^g, \mathbf{M}^p)}{\textrm{TP}(\mathbf{M}^g, \mathbf{M}^p)+\textrm{FP}(\mathbf{M}^g, \mathbf{M}^p)+\textrm{FN}(\mathbf{M}^g, \mathbf{M}^p)},
    \label{eq:jac}
\end{align}
where $\textrm{TP}$ denotes true positives, $\textrm{FP}$ denotes false positives, and $\textrm{FN}$ indicates false negatives. Most synthetic- and real-world graphs have multiple ground-truth explanations. For example, in the MUTAG dataset\cite{kazius2005derivation}, carbon rings with both NH$_2$ or NO$_2$ chemical groups are valid explanations for the GNN model to recognize a given molecule as mutagenic. For this reason, the accuracy metric must account for the possibility of multiple equally valid explanations existing for any given prediction. Hence, we define $\zeta$ as a set of all possible ground-truth explanations, where $|\zeta|=1$ for graphs having a unique explanation. Therefore, we calculate GEA as:
\begin{equation}
    \textrm{GEA}(\zeta, \mathbf{M}^p) = \max~\text{JAC}( \mathbf{M}^g,  \mathbf{M}^p)~~\forall \mathbf{M}^g \in \zeta.
    \label{eq:gt_faith}
\end{equation}
Here, we can calculate GEA using predicted node feature, node, or edge explanation masks. Finally, Equation~\ref{eq:jac} quantifies the accuracy between the ground-truth and predicted explanation masks. Higher values mean a predicted explanation is more likely to match the ground-truth explanation.

\xhdr{Graph explanation faithfulness (GEF)} We extend existing faithfulness metrics\cite{agarwal2021probing,sanchez2020evaluating} to quantify how faithful explanations are to an underlying GNN predictor. In particular, we obtain the prediction probability vector $\hat{y}_u$ using the GNN, \ie, $\hat{y}_u=f(\mathcal{S}_{u})$, and using the explanation, \ie, $\hat{y}_{u'}=f(\mathcal{S}_{u'})$, where we generate a masked subgraph $\mathcal{S}_{u'}$ by only keeping the original values of the top-$k$ features identified by an explanation, and get their respective predictions $\hat{y}_{u'}$. Finally, we compute the graph explanation unfaithfulness metric as:
\begin{equation}
    \text{GEF}(f, \mathcal{S}_u, \mathcal{S}_u') = 1 - \exp^{-\text{KL}(f(\mathcal{S}_u) || f(\mathcal{S}_u'))},
    \label{eq:pred_faith}
\end{equation}
where Kullback-Leibler (KL) divergence score quantifies the distance between two probability distributions, and the ``$||$" operator indicates statistical divergence measure. Note that Equation~\ref{eq:pred_faith} is a measure of the unfaithfulness of the explanation. So, higher values indicate a higher degree
of unfaithfulness.

\xhdr{Graph explanation stability (GES)} Formally, an explanation is defined to be stable if the explanation for a given graph and its perturbed counterpart (generated by making infinitesimally small perturbations to the node feature vector and associated edges) are similar\cite{agarwal2021probing,yuan2022explainability}. We measure graph explanation stability w.r.t. the changes in the model behavior. In addition to similar output labels for $\mathcal{S}_{u}$ and the perturbed $\mathcal{S}_{u'}$, we employ a second level of check where the difference between model behaviors for $\mathcal{S}_{u}$ and $\mathcal{S}_{u'}$ is bounded by an infinitesimal constant $\delta$, \ie, $||\mathcal{L}_{\mathcal{S}_{u}}{-}\mathcal{L}_{\mathcal{S}_{u'}}||_{p}\leq \delta$. Here, $\mathcal{L}(\cdot)$ refers to any form of model knowledge like output logits $\hat{y}_u$ or embeddings $\mathbf{z}_u$. We compute graph explanation instability as:
\begin{equation}
    \text{GES}(\mathbf{M}^{p}_{\mathcal{S}_u},\mathbf{M}^{p}_{\mathcal{S}_{u'}}) = \max~ D(\mathbf{M}^{p}_{\mathcal{S}_u}, \mathbf{M}^{p}_{\mathcal{S}_{u'}}),~~~~~\forall \mathcal{S}_{u'}\in\mathcal{B}(\mathcal{S}_u)
    \label{eq:rel_stab}
\end{equation}
where $D(\cdot)$ represents the cosine distance metric, $\mathbf{M}^{p}_{\mathcal{S}_u}$ and $\mathbf{M}^{p}_{\mathcal{S}_{u'}}$ are the predicted explanation masks for $\mathcal{S}_{u}$ and $\mathcal{S}_{u'}$, and $\mathcal{B}(\cdot)$ represents a $\delta$-radius ball around $\mathcal{S}_u$ for which the model behavior is same. 
Equation~\ref{eq:rel_stab} is a measure of instability, and higher values indicate a higher degree of instability.

\xhdr{Counterfactual fairness mismatch} To measure counterfactual fairness\cite{agarwal2021probing}, we verify if the
explanations corresponding to $\mathcal{S}_u$ and its counterfactual counterpart (where the protected node feature is modified) are similar (dissimilar) if the underlying model predictions are similar (dissimilar). We calculate counterfactual fairness mismatch as:
\begin{equation}
    \text{GECF}(\mathbf{M}^{p},\mathbf{M}^{p}_{s}) = D(\mathbf{M}^{p}, \mathbf{M}^{p}_{s}),
    \label{eq:counter}
\end{equation}
where, $\mathbf{M}^p$ and $\mathbf{M}^p_{s}$ are the predicted explanation mask for $\mathcal{S}_u$ and for the counterfactual counterpart of $\mathcal{S}_u$. Note that we expect a lower GECF score for graphs having weakly-unfair ground-truth explanations because explanations are similar for both original and counterfactual graphs, whereas, for graphs with strongly-unfair ground-truth explanations, we expect a higher GECF score as explanations change when we modify the protected attribute.

\xhdr{Group fairness mismatch} We measure group fairness mismatch\cite{agarwal2021probing} as:
\begin{equation}
    \text{GEGF}(\mathbf{\hat{y}}_{\mathcal{K}}, \mathbf{\hat{y}}_{\mathcal{K}}^{\mathbf{E}_u}) = |\text{SP}(\mathbf{\hat{y}}_{\mathcal{K}}) - \text{SP}(\mathbf{\hat{y}}_{\mathcal{K}}^{\mathbf{E}_u})~|,
    \label{eq:group_fair}
\end{equation}
where $\mathbf{\hat{y}}_{\mathcal{K}}$ and $\mathbf{\hat{y}}_{\mathcal{K}}^{\mathbf{E}_u}$ are predictions for a set of $\cK$ graphs using the original and the essential features identified by an explanation, respectively, and SP is the statistical parity. Finally, Equation~\ref{eq:group_fair} is a measure of group fairness mismatch of an explanation where higher values indicate that the explanation is not preserving group fairness.

\subsection*{Evaluation and analysis of GNN explainability methods}
Next, we discuss experimental results that answer critical questions concerning synthetic and real-world graphs and different ground-truth explanations.

\xhdr{Benchmarking GNN explainers on synthetic and real-world graphs} 
We evaluate the performance of GNN explainers on a collection of synthetically generated graphs with various properties and molecular datasets using metrics described in the experimental setup. Results in Tables~\ref{tab:benchmark_node}-\ref{tab:benchmark_graph} show that, while no explanation method performs well across all properties, across different \datagen node classification datasets (Table~\ref{tab:data-graph}), SubgraphX outperforms other methods on average. In particular, SubgraphX generates 145.95\% more accurate and 64.80\% less unfaithful explanations than other GNN explanation methods. Gradient-based methods, such as GradCam and GuidedBP, perform the next best of all methods, with Grad producing the second-lowest unfaithfulness score and GradCAM achieving the second-highest explanation accuracy score. PGExplainer generates the least unstable explanations---35.35\% less unstable explanations than the average instability across other GNN explainers. In summary, results of Table~\ref{tab:benchmark_node}-\ref{tab:benchmark_graph} show that i) node explanation masks are more reliable than edge- and node feature explanation masks and ii) state-of-the-art GNN explainers achieve better faithfulness for synthetic graph datasets as compared to real-world graphs.

\begin{figure}[ht]
    \centering
    \includegraphics[width=0.72\textwidth]{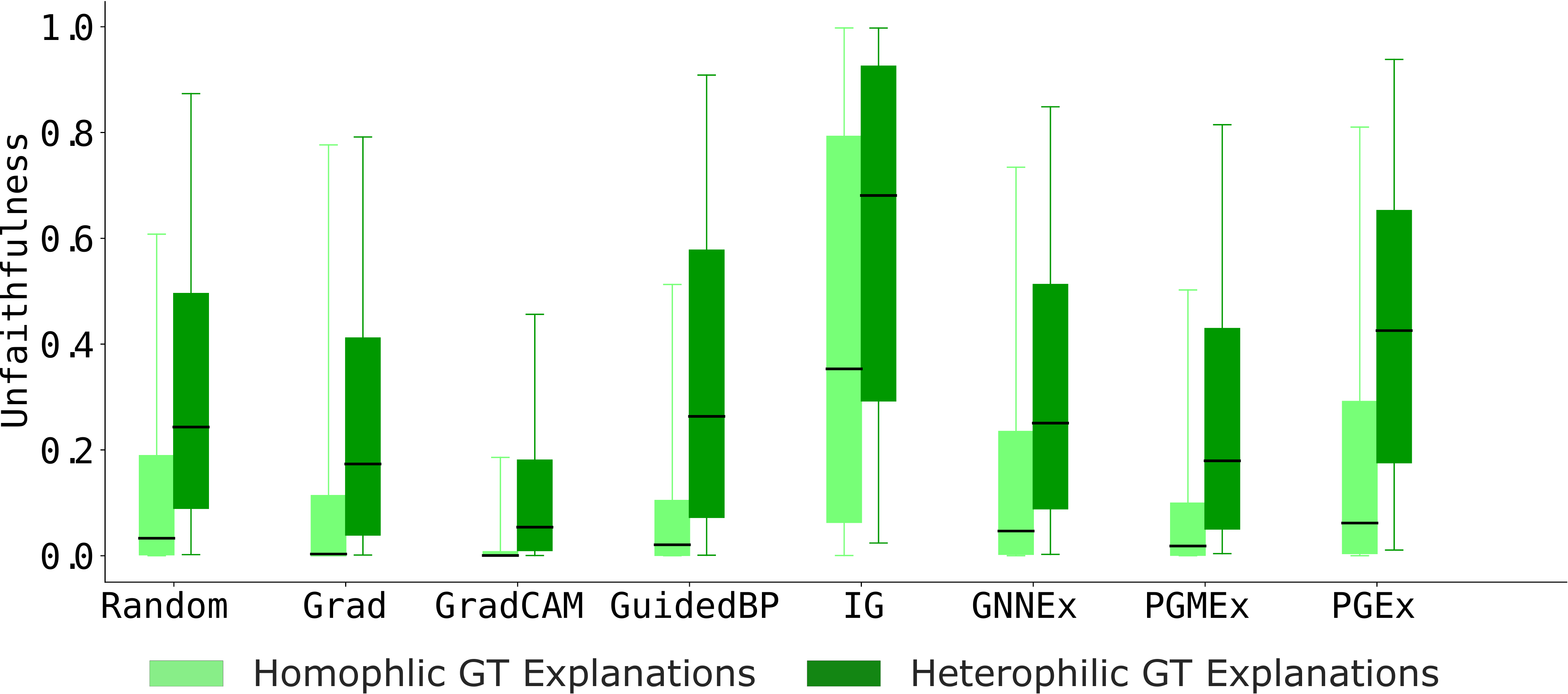}
    \caption{Unfaithfulness scores across eight GNN explainers on \sghetero graph dataset consisting of either homophilic or heterophilic ground-truth (GT) explanations. GNN explainers produce more faithful explanations (lower GEF scores) on homophilic graphs than heterophilic graphs, revealing an important limitation of existing GNN explainers.
    }
    \label{fig:homophiliy}
\end{figure}

\xhdr{Analyzing homophilic vs. heterophilic ground-truth explanations}
We compare GNN explainers by generating explanations on GNN models trained on homophilic and heterophilic graphs generated using the \sghetero generator. Then, we compute the graph explanation unfaithfulness scores of output explanations generated using state-of-the-art GNN explainers. We find that GNN explainers produce 55.98\% more faithful explanations when ground-truth explanations are homophilic than when ground-truth explanations are heterophilic (\ie, low unfaithfulness scores for light green bars in Figure~\ref{fig:homophiliy}). These results reveal an important gap in existing GNN explainers. Namely, existing GNN explainers fail to perform well on diverse graph types, such as homophilic, heterophilic and attributed graphs. This observation, enabled by the flexibility of \datagen generator, highlights an opportunity for future algorithmic innovation in GNN explainability.

\begin{figure}[ht]
    \centering
    \includegraphics[width=0.72\textwidth]{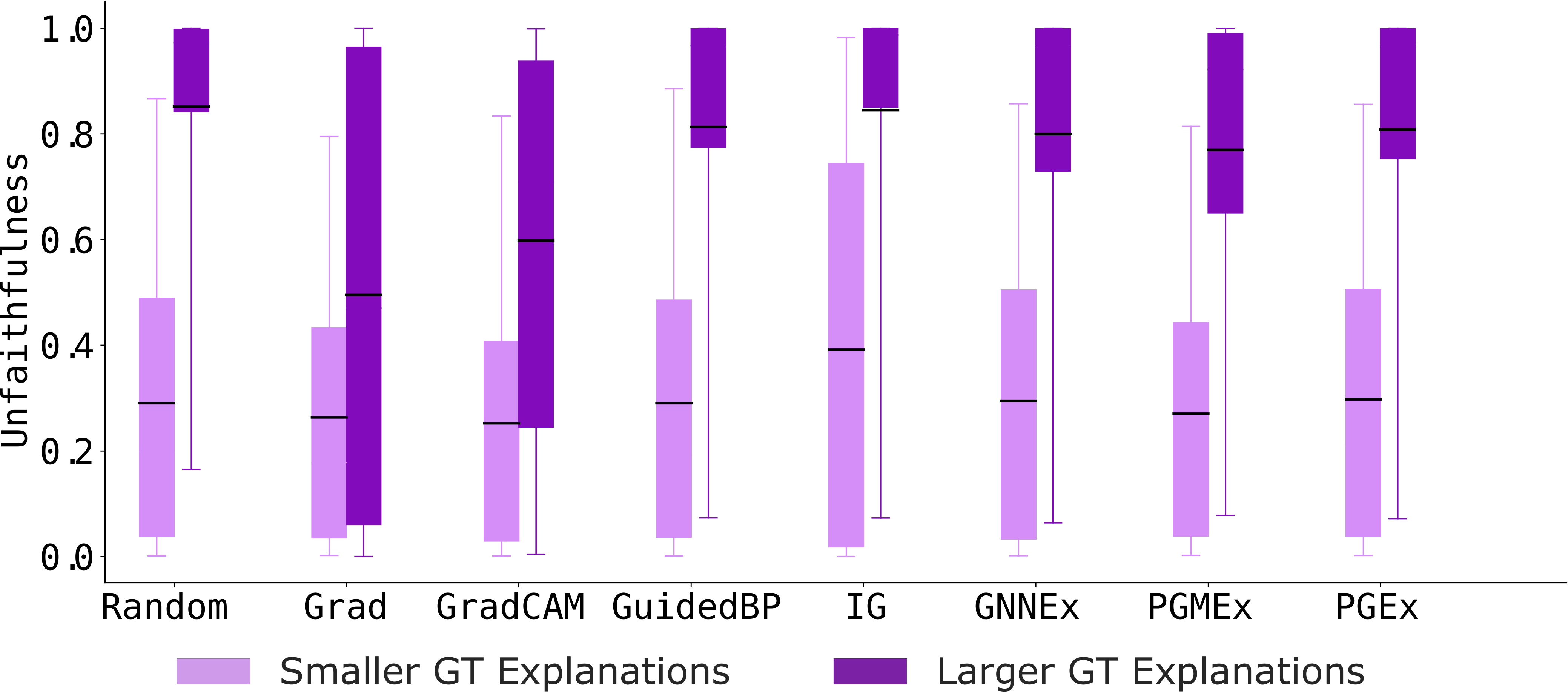}
    \caption{Unfaithfulness scores across eight GNN explainers on \sgsmallex graph dataset with smaller (triangle shapes) or (house shapes) ground-truth (GT) explanations. Results show that GNN explainers produce more faithful explanations (lower GEF scores) on graphs with smaller GT explanations than on graphs with larger GT explanations.
    }
    \label{fig:motifs}
\end{figure}

\xhdr{Analyzing the reliability of graph explainers to smaller vs. larger ground-truth explanations} 
Next, we examine the reliability of GNN explainers when used to predict explanations for models trained on graphs generated using the \sgsmallex graph generator. Results in Figure~\ref{fig:motifs} show that explanations from existing GNN explainers are faithful (\ie, lower GEF scores) to the underlying GNN models when ground-truth explanations are smaller, \ie, $\cS{=}$`triangle'. On average, across all 8 GNN explainers, we find that existing GNN explainers are highly unfaithful to graphs with large ground-truth explanations with an average GEF score of 0.7476. Further, we observe that explanations generated on `triangular' (smaller) ground-truth explanations are 59.98\% less unfaithful than explanations for `house' (larger) ground-truth explanations (\ie, low unfaithfulness scores for light purple bars in Figure~\ref{fig:motifs}). However, the Grad explainer, on average, achieves 9.33\% lower unfaithfulness on large ground-truth explanations compared to other explanation methods. This limited behavior of existing GNN explainers has not been previously known and highlights an urgent need for additional analysis of GNN explainers.

\begin{figure}[ht]
    \centering
    \includegraphics[width=0.72\textwidth]{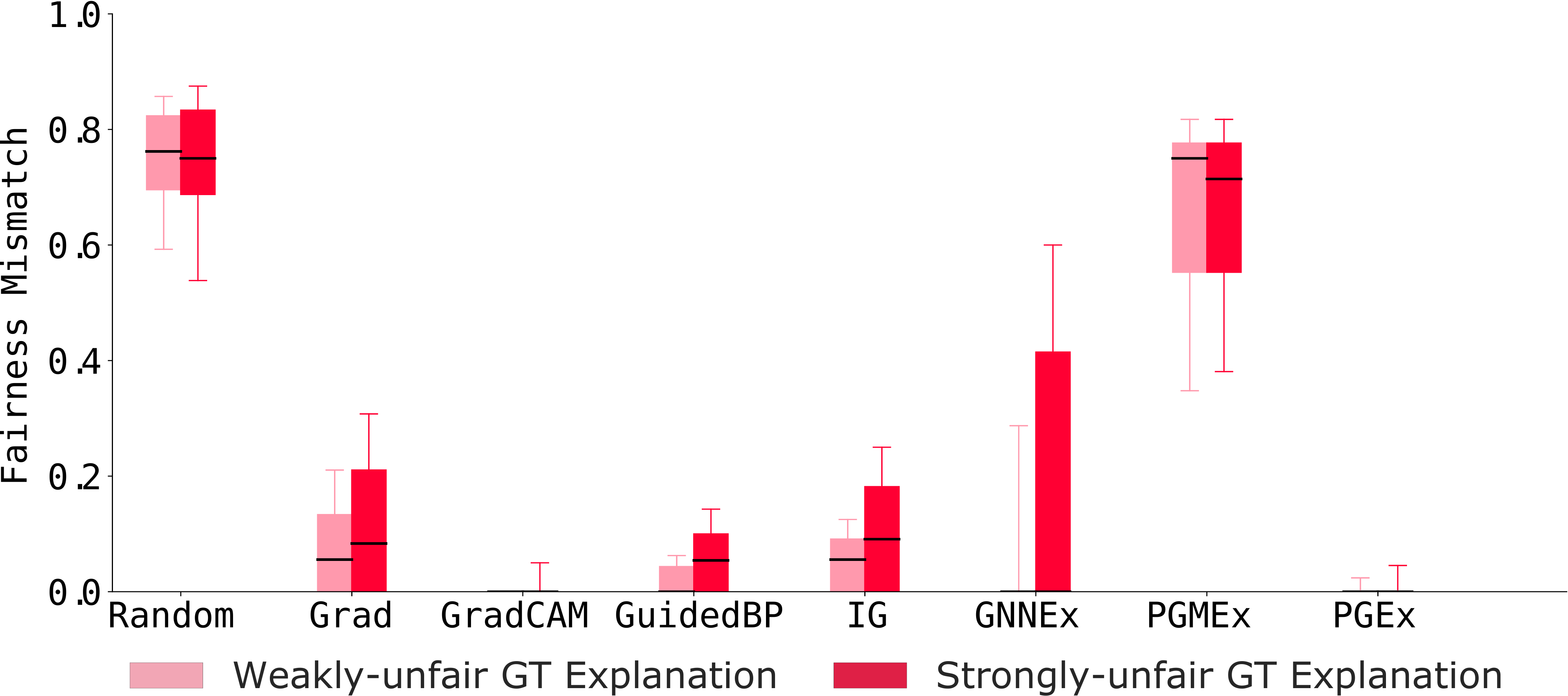}
    \caption{Counterfactual fairness mismatch scores across eight GNN explainers on \sgunfair graph dataset with weakly-unfair or strongly-unfair ground-truth (GT) explanations. Results show that explanations produced on graphs with strongly-unfair ground-truth explanations do not preserve fairness and are sensitive to flipping/modifying the protected node feature.
    }
    \label{fig:bias}
\end{figure}

\xhdr{Examining fair vs. unfair ground-truth explanations} 
To measure the fairness of predicted explanations, we train GNN models on \sgunfair, which generates graphs with controllable fairness properties. Next, we compute the average GECF and GEGF values for predicted explanations from eight GNN explainers. The fairness results in Figure~\ref{fig:bias} show that GNN explainers do not preserve counterfactual fairness and are highly prone to producing unfair explanations. We note that for weakly-unfair ground-truth explanations (light red in Figure~\ref{fig:bias}), explanations $\mathbf{M}^p$ should not change as the label-generating process is independent of the protected attribute. Still, we observe high GECF scores for most explanation methods. For strongly-unfair ground-truth explanations, we find that explanations from most GNN explainers fail to capture (\ie, low GECF scores for dark red bars in Figure~\ref{fig:bias}) the unfairness enforced using the protected attribute and generate similar explanations even when we flip/modify the respective protected attribute. We see that GradCAM and PGEx explanations outperform other GNN explainers in preserving counterfactual explanations for weakly-unfair ground-truth explanations. In contrast, the PGMEx explainer preserves counterfactual fairness better than other explanation methods on strongly-unfair ground truth explanations. Our results highlight the importance of studying fairness in XAI as they can enhance a user’s confidence in the model and assist in detecting and correcting unwanted bias.

\begin{figure}[ht]
    \centering
    \includegraphics[width=0.72\textwidth]{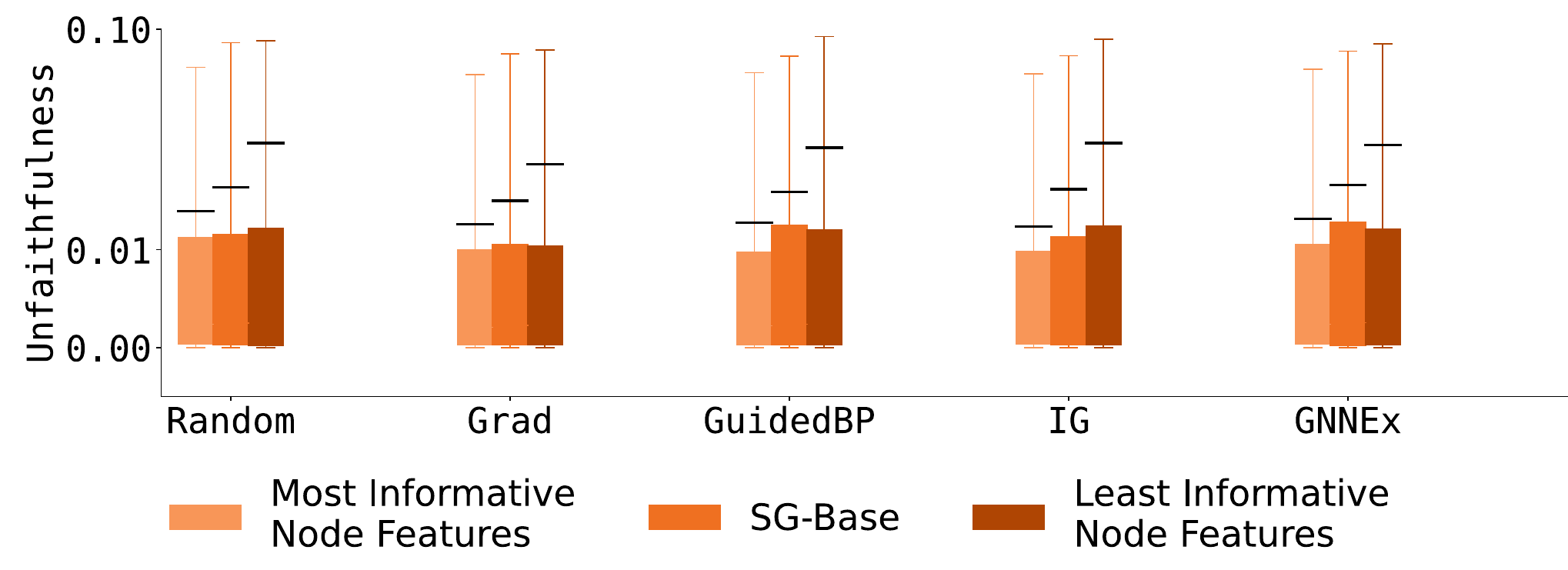}
    \caption{Unfaithfulness scores across five GNN explainers that produce node feature explanations. Every GNN explainer is evaluated on three datasets whose network topology is equivalent to \sgbase and by varying the ratio between informative and redundant node features: most informative node features, control node features, and least informative node features. Results show that across all explainers, unfaithfulness decreases as the proportion of informative to redundant features increases, with explainers trained on the graph with the most informative node features having consistently lower unfaithfulness scores than explainers trained on graphs with the least informative node features.}
    \label{fig:feature_exp}
\end{figure}

\xhdr{Faithfulness shift with varying degrees of node feature information}
Using \datagen's support for node features and ground-truth explanations on those features, we evaluate explainers that generate explanations for node features. Results for node feature explanations on \sgbase are given in Table \ref{tab:benchmark_feat}. In addition, we explore the performance of explainers under varying proportions of informative node features. Informative node features, defined in the \datagen construction (Algorithm \ref{alg:shapegraph}), are node features correlated with the label of a given node, as opposed to redundant features, which are sampled randomly from a Gaussian distribution. Figure \ref{fig:feature_exp} shows the results of experiments on three datasets, \sgmoreinform, \sgbase, and \sglessinform. All datasets have similar graph topology, but \sgmoreinform has a higher proportion of informative features while \sglessinform has a lower proportion of these informative features. \sgbase is used as a baseline with a proportion of informative features greater than \sglessinform but less than \sgmoreinform. There are minimal differences between explainers' faithfulness  across datasets, however, unfaithfulness tends to increase with fewer informative node features. As seen in Table \ref{tab:benchmark_feat}, the Gradient explainer shows the best faithfulness score across all datasets for node feature explanation. Still, this faithfulness is relatively weak, only 0.001 better than random explanation. These results show that the faithfulness of node feature explanations worsens under sparse node feature signals.

\begin{figure}[ht]
    \centering
    \includegraphics[width=0.90\textwidth]{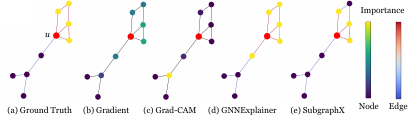}
    \caption{Visualization of four  explainers from the G-XAI Bench library on the BA-Shapes dataset. The visualization is for explaining the prediction of node $u$. We show the $L$+1-hop around node $u$, where $L$ is the number of layers of the GNN model predicting on the dataset. Two color bars indicate the intensity of attribution scores for the node and edge explanations. Note that edge importance is not defined for every method, so edges are set to black to indicate that the method does not provide edge scores. Visualization tools are a native part of the \name package, including user-friendly functions \texttt{graphxai.Explanation.visualize\_node} and \texttt{graphxai.Explanation.visualize\_graph} to visualize GNN explanations. The visualization tools in \name allow users to compare the explanations of different GNN explainers, such as gradient-based methods (Gradient and Grad-CAM) and perturbation-based methods (GNNExplainer and SubgraphX).
    }
    \label{fig:viz}
\end{figure}

\begin{figure}
    \centering
    \includegraphics[width=\textwidth]{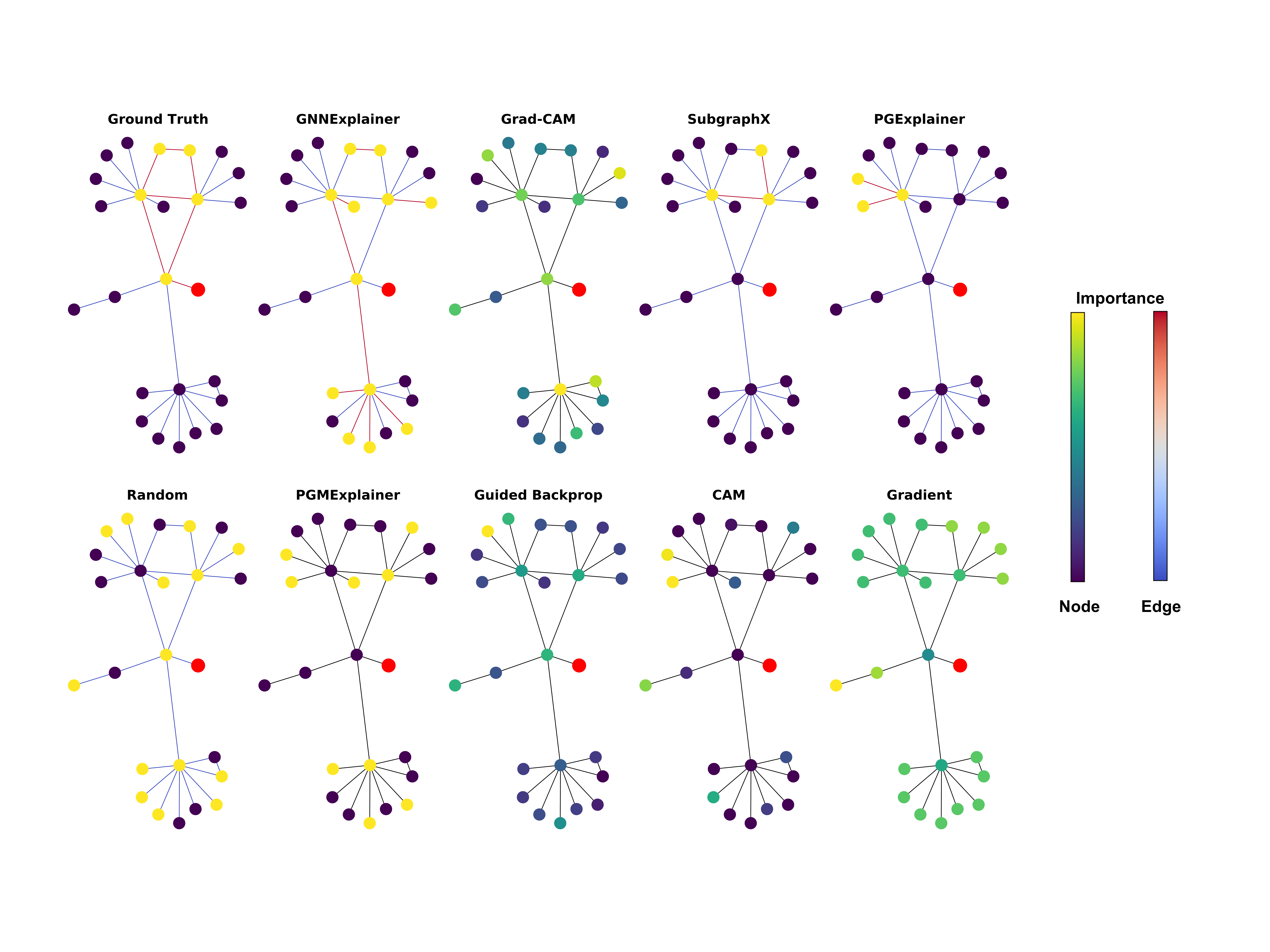}
    \caption{Example of a particularly challenging example in a \datagen dataset. All explanation methods that output node-wise importance scores are shown, including the ground-truth explanation at the top of the figure. Importance and edge scores are highlighted by relative value across each explanation method, as shown by the scales at right in the figure. The central node, \textit{i.e.}, the node being classified in this example, is shown in red on each subgraph. Visualizations are generated by \texttt{graphxai.Explanation.visualize\_node}, a function native to the \texttt{graphxai} package. Some explainers can capture portions of the ground-truth explanation, such as SubgraphX and GNNExplainer, but others attribute no importance to the ground-truth shape, such as CAM and Gradient.}
    \label{fig:shapegraph_viz}
\end{figure}

\xhdr{Visualization results} \name provides functions that visualize explanations produced by GNN explainability methods. Users can compare both node- and graph-level explanations. In addition, function implementations for visualization are parameterized, allowing users to change colors and weight interpretation. Functions are compatible with \texttt{matplotlib} \cite{hunter2007matplotlib} and \texttt{networkx} \cite{networkx}. Visualizations are generated by \texttt{graphxai.Explanation.visualize\_node} for node-level explanations and \texttt{graphxai.Explanation.visualize\_graph} functions for graph-level explanations. In Figure~\ref{fig:viz}, we show the output explanation from four different GNN explainers as produced by our visualization function. Figure \ref{fig:shapegraph_viz} shows example outputs from multiple explainers in the \name package on a \datagen-generated dataset.

\clearpage

\section*{Discussion}
\name provides a general-purpose framework to evaluate GNN explanations produced by state-of-the-art GNN explanation methods. \name provides data loaders, data processing functions, visualizers, real-world graph datasets with ground-truth explanations, and evaluation metrics to benchmark the quality of GNN explanations. \name introduces a novel and flexible synthetic dataset generator called \datagen to automatically generate benchmark datasets and corresponding ground truth explanations robust against known pitfalls of GNN explainability methods. Our experimental results show that existing GNN explainers perform well on graphs with homophilic ground-truth explanations but perform considerably worse on heterophilic and attributed graphs. Across multiple graph datasets and types of downstream prediction tasks, we show that existing GNN explanation methods fail on graphs with larger ground-truth explanations and cannot generate explanations that preserve the fairness properties of the underlying GNN model. In addition, GNN explainers tend to underperform on sparse node feature signals compared to more densely informative node features. These findings indicate the need for methodological innovation and a thorough analysis of future GNN explainability methods across performance dimensions. 

\name provides a flexible framework for evaluating GNN explanation methods and promotes reproducible and transparent research. We maintain \name as a centralized library for evaluating GNN explanation methods and plan to add newer datasets, explanation methods, diverse evaluation metrics, and visualization features to our existing framework. In the current version of \name, we mostly employ real-world molecular chemistry datasets as the need for model understanding is motivated by experimental evaluation of model predictions in the laboratory, and it includes a wide variety of graph sizes (ranging from 1,768 to 12,000 instances in the dataset), node feature dimensions (ranging from 13 to 27 dimensions), and class imbalance ratios. In addition to the scale-free \datagen dataset generator, we will include other random graph model generators in the next \name release to support benchmarking of GNN explainability methods on other graph types. Our evaluation metrics can be also extended to explanations from self-explaining GNNs, \eg, GraphMASK \cite{schlichtkrull2020interpreting} identifies edges at each layer of the GNN during training that can be ignored without affecting the output model predictions. In general, self-explaining GNNs also return a set of edge masks as an output explanation for a GNN prediction that can be converted to edge importance scores for computing GraphXAI metrics. We anticipate that \name can help algorithm developers and practitioners in graph representation learning develop and evaluate principled explainability techniques. 

\clearpage

% !TEX root = main.tex
\section*{Methods}
\datagen is a key component of \name and serves as a synthetic dataset generator of XAI-ready graph datasets. It is founded in graph theory and designed to address the pitfalls (see Introduction) of existing graph datasets in the broad area of explainable AI. As such, \datagen can facilitate the development, analysis, and evaluation of GNN explainability methods (see Results). We proceed with the description of \datagen data generator.

\subsection*{Notation}
\xhdr{Graphs} Let $\mathcal{G} = (\mathcal{V}_{\mathcal{G}}, \mathcal{E}_{\mathcal{G}}, \mathbf{X}_{\mathcal{G}})$ denote an undirected graph comprising of a set of nodes $\mathcal{V}_{\mathcal{G}}$ and a set of edges $\mathcal{E}_{\mathcal{G}}$. Let $\mathbf{X}_{\mathcal{G}}{=}\{\mathbf{x}_{1}, \mathbf{x}_{2}, \dots, \mathbf{x}_{N}\}$ denote the set of node feature vectors for all nodes in $\mathcal{V}_{\mathcal{G}}$, where $\mathbf{x}_{v} \in \mathbf{X}_{\mathcal{G}}$ is an $d$-dimensional vector which captures the attribute values of a node $v$ and $N{=}|\mathcal{V}_{\mathcal{G}}|$ denotes the number of nodes in the graph. Let $\mathbf{A} \in \mathbb{R}^{N \times N}$ be the graph adjacency matrix where element $\mathbf{A}_{uv}=1$ if there exists an edge $e \in \mathcal{E}_{\mathcal{G}}$ between nodes $u$ and $v$ and $\mathbf{A}_{uv}=0$ otherwise. We use ${\mathcal{N}}_u$ to denote the set of immediate neighbors of node $u$, \ie, ${\mathcal{N}}_{u}{=}\{ v \in \mathcal{V}_{\mathcal{G}} | A_{uv}=1 \}$. Finally, the function $\text{deg}: \mathcal{V}_{\mathcal{G}} \mapsto \mathbb{Z}_{\geq0}$ is defined as $\text{deg}(u) = | \mathcal{N}_{u} |$ and outputs the degree of a node $u \in \mathcal{V}_{\mathcal{G}}$.

\xhdr{Graph neural networks} Most GNNs can be formulated as message passing networks\cite{wu2020comprehensive} using three operators: \textsc{Msg}, \textsc{Agg}, and \textsc{Upd}. In a $L$-layer GNN, these operators are recursively applied on $\mathcal{G}$, specifying how neural messages (\ie embeddings) are exchanged between nodes, aggregated, and transformed to arrive at node representations in the last layer of transformations. Formally, a message between a pair of nodes $(u, v)$ in layer $l$ is defined as a function of hidden representations of nodes $\mathbf{h}_{u}^{l-1}$ and $\mathbf{h}_{v}^{l-1}$ from the previous layer: $\mathbf{m}_{uv}^l = \textsc{Msg}(\mathbf{h}_{u}^{l-1}, \mathbf{h}_{v}^{l-1})$. In \textsc{Agg}, messages from all nodes $v\in\mathcal{N}_{u}$ are aggregated as:~$\mathbf{m}_{u}^{l}=\textsc{Agg}(\mathbf{m}_{uv}^{l} | v\in{\mathcal{N}}_u)$. In \textsc{Upd}, the aggregated message $\mathbf{m}_{u}^{l}$ is combined with $\mathbf{h}_u^{l-1}$ to produce $u$'s representation for layer $l$  as $\mathbf{h}_{u}^{l}=\textsc{Upd}(\mathbf{m}_{u}^{l}, \mathbf{h}_{u}^{l-1})$. Final node representation $\mathbf{z}_u = \mathbf{h}_{u}^{L}$ is the output of the last layer. Lastly, let $f$ denote a downstream GNN classification model that maps the node representation $\mathbf{z}_{u}$ to a softmax prediction vector $\hat{y}_{u}\in\mathbb{R}^{C}$, where $C$ is the total number of labels.

\xhdr{GNN explainability methods} 
Given the prediction $f(u)$ for node $u$ made by a GNN model, a GNN explainer $\mathcal{O}$ outputs an explanation mask $\mathbf{M}^{p}$ that provides an explanation of $f(u)$. These explanations can be given with respect to node attributes $\mathbf{M}_{a}\in\mathbb{R}^{d}$, nodes $\mathbf{M}_{n}\in\mathbb{R}^{N}$, or edges $\mathbf{M}_{e}\in\mathbb{R}^{N\times N}$, depending on specific GNN explainer, such as GNNExplainer\cite{ying2019gnnexplainer}, PGExplainer\cite{luo2020parameterized}, and SubgraphX\cite{yuan2021explainability}. For all explanation methods, we use a graph masking function $\mathrm{MASK}$ that outputs a new graph $\mathcal{G'}=(\mathcal{V}'_{\mathcal{G}'}, \mathcal{E}'_{\mathcal{G}'}, \mathcal{X}'_{\mathcal{G}'})$ by performing element-wise multiplication operation between the masks $(\mathbf{M}_{a},\mathbf{M}_{n},\mathbf{M}_{e})$ and their respective attributes in the original graph $\mathcal{G}$, \eg $\mathbf{A'}=\mathbf{A} \odot \mathbf{M}_{e}$. Finally, we denote the ground-truth explanation mask as  $\mathbf{M}^{g}$ that is used to evaluate the performance of GNN explainers.

\subsection*{\datagen dataset generator}
\label{sec:shapegraph}
We propose a novel and flexible synthetic dataset generator called \datagen that can automatically generate a variety of benchmark datasets (\textit{e.g.}, varying graph sizes, degree distributions, homophilic vs. heterophilic graphs) accompanied by ground-truth explanations. Furthermore, the flexibility to generate diverse synthetic datasets and corresponding ground-truth explanations allows us to mimic the data generated by various real-world applications. \datagen is a generator of XAI-ready graph datasets supported by graph theory and is particularly suitable for benchmarking GNN explainers and studying their limitations.

\xhdr{Flexible parameterization of \datagen} 
\datagen has tunable parameters for data generation. By varying these parameters, \datagen can generate diverse types of graphs, including graphs with varying degrees of class imbalance and graph sizes. Formally, a graph is generated as: $\mathcal{G}=$\datagen$(\mathcal{S}, N_{s}, p, n_{s}, K, n_{f}, n_{\textrm{i}}, s_{f}, c_{f}, \phi, \eta, L)$, where:

% 
% \begin{itemize}
\begin{description}[font=$\bullet$~\normalfont\scshape\color{red!50!black}]
    \item $\mathcal{S}$ is the motif, defined as a subgraph, to be planted within the graph. 
    \item $N_{s}$ denotes the number of subgraphs used in the initial graph generation process.
    \item $p$ represents the connection probability between two subgraphs and controls the average shortest path length for all possible pairs of nodes. Ideally, a smaller $p$ value for larger $N_{s}$ is preferred to avoid low average path length and, therefore, the poor performance of GNNs.
    \item $n_{s}$ is the expected size of each subgraph in the \datagen generation procedure. For a fixed $\mathcal{S}$ shape, large $n_{s}$ values produce graphs with long-tailed degree distributions. Note that the expected total number of nodes in the generated graph $\mathcal{G}$ is $N \times n_{s}$.
    \item $K$ is the number of distinct classes defined in the graph downstream task.
    \item $n_{f}$ represents the number of dimensions for node features in the generated graph.
    \item $n_{\textrm{i}}$ is the number of informative node features. These are features correlated to the node labels instead of randomly generated non-informative features. The indices for the informative features define the ground-truth explanation mask for node features in the final \datagen instance. In general, larger $n_{\textrm{i}}$ results in an easier classification and explanation task, as it increases the node feature-level ground-truth explanation size.
    \item $s_{f}$ is defined as the class separation factor that represents the strength of discrimination of class labels between features for each node. Higher $s_{f}$ corresponds to a stronger signal in the node features, \textit{i.e.}, if a classifier is trained only on the node features, a higher $s_{f}$ value would result in an easier classification task.
    \item $c_{f}$ is the number of clusters per class. A larger $c_{f}$ value increases the difficulty of the classification task with respect to node features.
    \item $\phi \in [0, 1]$ is the protected feature noise factor that controls the strength of correlation $r$ between the protected features and the node labels. This value corresponds to the probability of ``flipping'' the protected feature with respect to the node's label. For instance, $\phi{=}0.5$ results in zero correlation ($r{=}0$) between the protected feature and the label (\textit{i.e.} complete fairness), $\phi{=}0$ results in a positive correlation ($r{=}1$), and $\phi{=}1$ results in a negative correlation ($r{=}-1$) between the label and the protected feature.
    \item $\eta$ is the homophily coefficient that controls the strength of homophily or heterophily in the graph. Positive values ($\eta{>}0$) produce a homophilic graph while negative values ($\eta{<}0$) produce a heterophilic graph. 
    \item $L$ is the number of layers in the GNN predictor corresponding to the GNN's receptive field. For the purposes of $\datagen$, $L$ is used to define the size of the GNN's receptive field and thus the size of the ground-truth explanation generated for each node.
% \end{itemize}
\end{description}

\noindent This wide array of parameters for \datagen allows for the generation of graph instances with vastly differing properties. 

\xhdr{Generating graph structure}
Figure \ref{fig:shapegraph} summarizes the process to generate a graph $\mathcal{G} = (\mathcal{V}_{\mathcal{G}}$, $\mathcal{E}_{\mathcal{G}}, \mathbf{X}_{\mathcal{G}})$. \datagen generates $N_{s}$ subgraphs that exhibit the preferential attachment property\cite{barabasi1999emergence}, which occurs in many real-world graphs. Each subgraph is first given a motif, \ie, a subgraph $\mathcal{S} = (\mathcal{V}_{\mathcal{S}}, \mathcal{E}_{\mathcal{S}}, \mathbf{X}_{\mathcal{S}})$. A preferential attachment algorithm is then performed on base structure $\mathcal{S}$, adding $n'$ ($n' \sim \text{Poisson}(\lambda = n_s - |\mathcal{V}_{\mathcal{S}}|)$) nodes by creating edges to nodes in $\mathcal{V}_{\mathcal{S}}$. The Poisson distribution is used for determining the sizes of each subgraph used in the generation process, with $\lambda = n_s - |\mathcal{V}_{\mathcal{S}}|$, the difference between the number of nodes in the motif and the expected subgraph size $n_s$. After creating a list of randomly-generated subgraphs $\mathbf{S} = \{ \mathcal{S}_1, \dots, \mathcal{S}_N\}$, edges are created to connect subgraphs, creating the structure of an instance of \datagen. Subgraph connections and local subgraph construction is performed in such a way that each node in the final graph $\mathcal{G}$ has between 1 and $K$ motifs within its neighborhood, \ie, $\Big| \bigcup_{i=1}^{N}\mathcal{V}_{\mathcal{S}_i} \cap \mathcal{N}_{v} \Big|\ {\in}\ \{1, 2, ..., K\}$ for any $v$ and $\mathcal{S}_i$. This naturally defines a classification task in the domain of $f$ to $\{0,1, ..., K{-}1\}$. More details on the \datagen structure generation can be found in Algorithm \ref{alg:shapegraphstructure}.

\xhdr{Generating labels for prediction}
A motif defined as a subgraph $\mathcal{S} = (\mathcal{V}_{\mathcal{S}}, \mathcal{E}_{\mathcal{S}}, \mathbf{X}_{\mathcal{S}})$ occurs randomly throughout $\mathcal{G}$, with the set $\mathbf{S} = \{ \mathcal{S}_{1}, \dots, \mathcal{S}_{N}\}$. The task on this graph is a motif detection problem, where each node has exactly 1, 2, or $K$ motifs in its $1$-hop neighborhood. A motif $\mathcal{S}_{i}$ is considered to be within the neighborhood of a node $v \in \mathcal{V}_{\mathcal{G}}$ if any node $s \in \mathcal{V}_{\mathcal{S}_i}$ is also in the neighborhood of $v$, \ie, if $|\mathcal{V}_{\mathcal{S}_i} \cap \mathcal{N}_{v}| > 0$. Therefore, the task that a GNN predictor needs to solve is defined by:
\begin{equation}
f(v \in \mathcal{V}_{\mathcal{G}}) = \sum\limits_{\mathcal{S}_i \in \mathbf{S}} \mathbbm{1}_{\mathcal{V}_{\mathcal{S}_i}}(\mathcal{N}_{v}) - 1,
\label{eq:shapegraph_f}
\end{equation}
where $\mathbbm{1}_{\mathcal{V}_{\mathcal{S}_i}}(\mathcal{N}_{v}) = 0$ if $|\mathcal{V}_{\mathcal{S}_i} \cap \mathcal{N}_{v}| = 0$ and $1$ otherwise.

\xhdr{Generating node feature vectors}
\datagen uses a latent variable model to create node feature vectors and associate them with network structure. This latent variable model is based on that developed by Guyon \cite{guyon2003design} for the MADELON dataset and implemented in Scikit-Learn's \texttt{make\_classification} function \cite{scikit-learn}. The latent variable model creates $n_{\text{i}}$ informative features for each node based on the node's generated label and also creates non-informative features as noise. Having non-informative/redundant features allows us to evaluate GNN explainers, such as GNNExplainer\cite{ying2019gnnexplainer}, that formulate explanations as node feature masks. More detail on node feature generation is given in Algorithm \ref{alg:shapegraph}.

\datagen can generate graphs with both homophilic and heterophilic ground-truth explanations. We optimize between homophily vs. heterophily by taking advantage of redundant node features, \ie, features that do not associate with the generated labels, and manipulate them appropriately based on a user-specified homophily parameter $\eta$. The magnitude of the $\eta$ parameter determines the degree of homophily/heterophily in the generated graph. The algorithm for node features is given in Algorithm~\ref{alg:optimizehomophily}. While not every node feature in the feature vector is optimized for homophily/heterophily indication, we experimentally verified the cosine similarity between node feature vectors produced by Algorithm~\ref{alg:optimizehomophily} reveals a strong homophily/heterophily pattern. Finally, \datagen can generate protected features to enable the study of fairness\cite{agarwal2021towards}. By controlling the value assignment for a selected discrete node feature, \datagen introduces bias between the protected feature and node labels. The biased feature is a proxy for a protected feature, such as gender or race. The procedure for generating node features is outlined in \textsc{\textbf{NodeFeatureVectors}} within Algorithm~\ref{alg:shapegraph}.

\xhdr{Generating ground-truth explanations}
In addition to generating ground-truth labels, \datagen has a unique capability to generate unique ground-truth explanations. To accommodate diverse types of GNN explainers, every ground-truth explanation in \datagen contains information on a) the importance of nodes, b) the importance of node features, and c) the importance of edges. This information is represented by three distinct masks defined over enclosing subgraphs of nodes $v \in \mathcal{V}_{\mathcal{G}}$, \textit{i.e}., the $L$-hop neighborhood around the node. We denote the enclosing subgraph of node $v \in \mathcal{V}_{\mathcal{G}}$ for a given GNN predictor with $L$ layers as: $\textsc{Sub}(v;L) = (\mathcal{V}_{\textsc{Sub}(v)}, \mathcal{E}_{\textsc{Sub}(v)},\mathbf{X}_{\textsc{Sub}(v)}) \subseteq \mathcal{G}$. Let motifs within this enclosing subgraph be: $\mathbf{S}_v = (\mathcal{V}_{\mathbf{S}_v}, \mathcal{E}_{\mathbf{S}_v}, \mathbf{X}_{\mathbf{S}_v}) = \mathbf{S} \cap \textsc{Sub}(v)$. Using this notation, we define ground-truth explanation masks:
% with $L$ indicating the number of layers in the GNN predictor

\xhdr{a) Node explanation mask} Nodes in $\textsc{Sub}(v)$ are assigned a value of 0 or 1 based on whether they are located within a motif or not. For any node $v_i \in \mathcal{V}_{\textsc{Sub}(v)}$, we set $\mathbbm{1}_{\mathcal{V}_{\mathbf{S}_v}}(v_i) =  1$ if $v_i \in \mathcal{V}_{\mathbf{S}_v}$ and 0 otherwise.
This function $\mathbbm{1}_{\mathcal{V}_{\mathbf{S}_v}}$ is applied to all nodes in the enclosing subgraph of $v$ to produce an importance score for each node, yielding the final mask as: $\mathbf{M}_{n}=\{\mathbbm{1}_{\mathcal{V}_{\mathbf{S}_v}}(u) | u \in \mathcal{V}_{\textsc{Sub}(v)}\}$.
    
\xhdr{b) Node feature explanation mask}
Each feature in $v$'s feature vector is labeled as 1 if it represents an informative feature and 0 if it is a random feature. This procedure produces a binary node feature mask for node $v$ as: $\mathbf{M}_{f}\in\{0, 1\}^{d}$.

\xhdr{c) Edge explanation mask}
To each $e = (v_i, v_j) \in \mathcal{E}_{\textsc{Sub}(v)}$ we assign a value of either 0 or 1 based on whether $e$ connects any two nodes in $\textsc{Sub}(v)$.
% in a motif or the source node $v$ to some node within a motif. 
The masking function is defined as follows:
\begin{equation}
    \mathbbm{1}_{\mathcal{E}_{\mathbf{S}_v}}(e) = 
    \begin{cases}
       0 & \text{if $v_i \notin (\mathcal{V}_{\mathbf{S}_v} \cup \{v\}) \lor v_j \notin (\mathcal{V}_{\mathbf{S}_v} \cup \{v\})$} \\
       1 & \text{if $v_i \in (\mathcal{V}_{\mathbf{S}_v} \cup \{v\}) \wedge v_j \in (\mathcal{V}_{\mathbf{S}_v} \cup \{v\})$} \\
    \end{cases}
    \label{eq:gtedgeassign}
\end{equation}
This function $\mathbbm{1}_{\mathcal{E}_{\mathbf{S}_v}}$ is applied to all edges $e \in \mathcal{E}_{\textsc{Sub}(v)}$ to produce ground-truth edge explanation as: $\mathbf{M}_e = \{ \mathbbm{1}_{\mathcal{E}_{\mathbf{S}_v}}(e) | e \in \mathcal{E}_{\textsc{Sub}(v)}\}$. The procedure to generate these ground-truth explanations is thoroughly described in Algorithm \ref{alg:shapegraph}.

\subsection*{Datasets in \name}
\label{sec:datasets}
We proceed with a detailed description of synthetic and real-world graph data resources included in \name.

\subsubsection*{Synthetic graphs}
The \datagen generator outlined in the Methods section is a dataset generator that can be used to generate any number of user-specified graphs. In \name, we provide a collection of pre-generated XAI-ready graphs with diverse properties that are readily available for analysis and experimentation. 

\xhdr{Base \datagen graphs (\sgbase)} We provide a base version of \datagen graphs. This instance of \datagen is homophilic, large, and contains house-shaped motifs for ground-truth explanations, formally described as $\mathcal{G}$ = \datagen ($\mathcal{S}{=}$`house', $N_{s}{=}1200$, $p{=}0.006$, $n_s{=}11$, $K{=}2$, $n_{{f}}{=}11$, $n_{{i}}{=}4$, $s_{{f}}{=}0.6$, $c_{{f}}{=}2$, $\phi{=}0.5$, $\eta{=}1$, $L{=}3$). The node features in this graph exhibit homophily, a property commonly found in social networks. With over 10,000 nodes, this graph also provides enough examples of ground-truth explanations for rigorous statistical evaluation of explainer performance. The house-shaped motifs follow one of the earliest synthetic graphs used to evaluate GNN explainers\cite{ying2019gnnexplainer}.

\xhdr{Homophilic and heterophilic \datagen graphs} GNN explainers are evaluated on homophilic graphs\cite{agarwal2021towards, mccallum2000automating, sen2008collective, wang2020microsoft} as low homophily levels in graphs can degrade the performance of GNN predictors\cite{zhu2020beyond,jin2021node}. To this end, there are no heterophilic graphs with ground-truth explanations in current GNN XAI literature despite such graphs being abundant in real-world applications\cite{zhu2020beyond}. To demonstrate the flexibility of the \datagen data generator, we use it to generate graphs with: i) homophilic ground-truth explanations (\sgbase) and ii) heterophilic ground-truth explanations (\sghetero), \textit{i.e.}, $\mathcal{G}$ = \datagen ($\mathcal{S}{=}$`house', $N_{s}{=}1200$, $p{=}0.006$, $n_s{=}11$, $K{=}2$, $n_{{f}}{=}11$, $n_{{i}}{=}4$, $s_{{f}}{=}0.6$, $c_{{f}}{=}2$, $\phi{=}0.5$, $\eta{=}-1$, $L{=}3$).

\xhdr{Weakly and strongly unfair \datagen graphs}  We utilize the \datagen data generator to generate graphs with controllable fairness properties, \textit{i.e.}, leverage \datagen to generate synthetic graphs with real-world fairness properties where we can enforce unfairness w.r.t. a given protected attribute. We use the \datagen to generate graphs with: i) weakly-unfair ground-truth explanations (\sgbase) and ii) strongly-unfair ground-truth explanations (\sgunfair) $\mathcal{G}$ = \datagen ($\mathcal{S}{=}$`house', $N_{s}{=}1200$, $p{=}0.006$, $n_s{=}11$, $K{=}2$, $n_{{f}}{=}11$, $n_{{i}}{=}4$, $s_{{f}}{=}0.6$, $c_{{f}}{=}2$, $\phi{=}0.75$, $\eta{=}1$, $L{=}3$). Here, for the first time, we generated unfair synthetic graphs that can serve as pseudo-ground-truth for quantifying whether current GNN explainers preserve counterfactual fairness.

\xhdr{Small and large \datagen explanations} We explore the faithfulness of explanations w.r.t. different ground-truth explanation sizes. This is important because a reliable explanation should identify important features independent of the explanation size. However, current data resources only provide graphs with smaller-size ground-truth explanations. Here, we use the \datagen data generator to generate graphs having i) smaller ground-truth explanations size (\sgsmallex) $\mathcal{G}$ = \datagen ($\mathcal{S}{=}$`triangle', $N_{s}{=}1200$, $p{=}0.006$, $n_s{=}12$, $K{=}2$, $n_{{f}}{=}11$, $n_{{i}}{=}4$, $s_{{f}}{=}0.5$, $c_{{f}}{=}2$, $\phi{=}0.5$, $\eta{=}1$, $L{=}3$) and ii) larger ground-truth explanations size (\sgbase), \ie house motifs.

\xhdr{Low and high proportions of salient features}
We examine the faithfulness of node feature masks produced by explainers under different levels of sparsity for class-associated signal in the node features. The feature generation procedure in \datagen specifies node feature parameters $n_{\textrm{i}}$, the number of informative features that are generated to correlate with node labels, and $n_{\textrm{f}}$, number of total node features. The remaining $n_{\textrm{f}} - n_{\textrm{i}}$ features are redundant features that are randomly distributed and have no correlation to the node label. Using an equivalent graph topology to \sgbase, we change the relative proportion of node features which are attributed to the label by adjusting $n_{\textrm{i}}$ and $n_{\textrm{f}}$. We create \sgmoreinform, a \datagen instance with a high proportion of ground-truth features to total features (8:11). Likewise, we create \sglessinform, a \datagen instance with a low proportion of ground-truth features to total features (4:21). This proportion in \sgbase falls in the middle of \sgmoreinform and \sglessinform instances with a proportion of ground-truth to total features of 4:11. Formally, we define \sgmoreinform as $\mathcal{G}$ = \datagen ($\mathcal{S}{=}$`house', $N_{s}{=}1200$, $p{=}0.006$, $n_s{=}11$, $K{=}2$, $n_{{f}}{=}11$, $n_{{i}}{=}8$, $s_{{f}}{=}0.6$, $c_{{f}}{=}2$, $\phi{=}0.5$, $\eta{=}1$, $L{=}3$) and \sglessinform as $\mathcal{G}$ = \datagen ($\mathcal{S}{=}$`house', $N_{s}{=}1200$, $p{=}0.006$, $n_s{=}11$, $K{=}2$, $n_{{f}}{=}21$, $n_{{i}}{=}4$, $s_{{f}}{=}0.6$, $c_{{f}}{=}2$, $\phi{=}0.5$, $\eta{=}1$, $L{=}3$).

\xhdr{BA-Shapes} In addition to \datagen, we provide a version of \textsc{BA-Shapes}\cite{ying2019gnnexplainer}, a synthetic graph data generator for node classification tasks. We start with a base Barabasi-Albert (BA)\cite{albert2002statistical} graph using $N$ nodes and a set of five-node ``house''-structured motifs $K$ randomly attached to nodes of the base graph. The final graph is perturbed by adding random edges. The nodes in the output graph are categorized into two classes corresponding to whether the node is in a house (1) or not in a house (0).

\subsubsection*{Real-world graphs}
We describe the real-world graph datasets with and without ground-truth explanations provided in \name. To this end, we provide data resources from crime forecasting, financial lending, and molecular chemistry and biology\cite{agarwal2021towards,sanchez2020evaluating,kazius2005derivation}. We consider these datasets as they are used to train GNNs for generating predictions in high-stakes downstream applications. In particular, we include chemical and biological datasets because they are used to identify whether an input graph (\textit{i.e.}, a molecular graph) contains a specific pattern (\textit{i.e.}, a chemical group with a specific property in the molecule). Knowledge about such patterns, which determine molecular properties, represents ground-truth explanations\cite{sanchez2020evaluating}. We provide a statistical description of real-world graphs in Tables~\ref{tab:data-graph}-\ref{tab:data-node}. Below, we discuss the details of each of the real-world datasets that we employ:

\xhdr{MUTAG} The MUTAG\cite{kazius2005derivation} dataset contains 1,768 graph molecules labeled into two different classes according to their mutagenic properties, \textit{i.e.}, effect on the Gram-negative bacterium \textit{S. Typhimuriuma}. Kazius et al.\cite{kazius2005derivation} identifies several toxicophores - motifs in the molecular graph - that correlate with mutagenicity. The dataset is trimmed from its original 4,337 graphs to 1,768, based on those whose labels directly correspond to the presence or absence of our chosen toxicophores: NH$_2$, NO$_2$, aliphatic halide, nitroso, and azo-type (terminology, as referred to in Kazius et al.\cite{kazius2005derivation}).

\xhdr{Alkanecarbonyl} The Alkane-Carbonyl\cite{sanchez2020evaluating} dataset contains 1,125 molecular graphs labeled into two classes where a positive sample indicates a molecule that contains an unbranched alkane and a carbonyl (C=O) functional group. The ground-truth explanations include any combinations of alkane and carbonyl functional groups within a given molecule.

\xhdr{Benzene} The Benzene\cite{sanchez2020evaluating} dataset contains 12,000 molecular graphs extracted from the ZINC15\cite{sterling2015zinc} database and labeled into two classes where the task is to identify whether a given molecule has a benzene ring or not. Naturally, the ground truth explanations are the nodes (atoms) comprising the benzene rings, and in the case of multiple benzenes, each of these benzene rings forms an explanation. 

\xhdr{Fluoride carbonyl} The Fluoride~Carbonyl\cite{sanchez2020evaluating} dataset contains 8,671 molecular graphs labeled into two classes where a positive sample indicates a molecule that contains a fluoride (F$^{-}$) and a carbonyl (C=O) functional group. The ground-truth explanations consist of combinations of fluoride atoms and carbonyl functional groups within a given molecule.

\xhdr{German credit} The German Credit\cite{agarwal2021towards} graph dataset contains 1,000 nodes representing clients in a German bank connected based on the similarity of their credit accounts. The dataset includes demographic and financial features like Gender, Residence, Age, Marital Status, Loan Amount, Credit History, and Loan Duration. Finally, the goal is to classify clients into good vs. bad credit risks.

\xhdr{Recidivism} The Recidivism\cite{agarwal2021towards} dataset includes samples of bail outcomes collected from multiple state courts in the USA between 1990-2009. It contains past criminal records, demographic attributes, and other demographic details of 18,876 defendants (nodes) who got released on bail at the U.S. state courts. Defendants are connected based on the similarity of past criminal records and demographics, and the goal is to classify defendants into bail vs. no bail.

\xhdr{{Credit defaulter}} The Credit defaulter\cite{agarwal2021towards} graph has 30,000 nodes representing individuals that we connected based on the similarity of their spending and payment patterns. The dataset contains applicant features like education, credit history, age, and features derived from their spending and payment patterns. The task is to predict whether an applicant will default on an upcoming credit card payment.

\clearpage

\paragraph{Data availability.} \name data resource~\cite{graphXAI22} is hosted on \href{https://dataverse.harvard.edu/dataset.xhtml?persistentId=doi:10.7910/DVN/KULOS8}{Harvard Dataverse} under a persistent identifier \url{https://doi.org/10.7910/DVN/KULOS8}. We have deposited different a number of \datagen-generated datasets and real-world graphs at this repository.

\paragraph{Code availability.} Project website for \name is at \url{https://zitniklab.hms.harvard.edu/projects/GraphXAI}. The code to reproduce results, documentation, and tutorials are available in \name's Github repository at \href{https://github.com/mims-harvard/GraphXAI}{https://github.com/mims-harvard/GraphXAI}. The repository contains Python scripts to generate, evaluate the explanation using performance metrics, and visualize the explanation. In addition, the repository contains information and Python scripts to build new versions of \name as the underlying primary resources get updated and new data become available.

\paragraph{Acknowledgements.} 
C.A., O.Q., and M.Z. gratefully acknowledge the support by NSF under Nos. IIS-2030459 and IIS-2033384, US Air Force Contract No. FA8702-15-D-0001, and awards from Harvard Data Science Initiative, Amazon Research, Bayer Early Excellence in Science, AstraZeneca Research, and Roche Alliance with Distinguished Scientists. H.L. was supported in part by NSF under Nos IIS-2008461 and IIS-2040989, and research awards from Google, JP Morgan, Amazon, Bayer, Harvard Data Science Initiative, and D3 Institute at Harvard. O.Q. was supported, in part, by Harvard Summer Institute in Biomedical Informatics. Any opinions, findings, conclusions or recommendations expressed in this material are those of the authors and do not necessarily reflect the views of the funders.

\paragraph{Author contributions.} 
C.A., O.Q., H.L., and M.Z. contributed new analytic tools and wrote the manuscript. C.A. and O.Q. retrieved, processed, and harmonized datasets. C.A. and O.Q. implemented the synthetic dataset generator and performed the analyses for technical validation of the new resource. M.Z. conceived the study. 

\paragraph{Competing interests.} 

The authors declare no competing interests.

\clearpage

% !TEX root = main.tex

\newgeometry{left=0.7in,right=0.7in}
\captionsetup{margin=0.1in}
\spacing{1}
\pagestyle{empty}

\begin{figure}[ht]
     \centering
     \begin{subfigure}[b]{0.45\textwidth}
         \centering
         \includegraphics[width=0.81\textwidth]{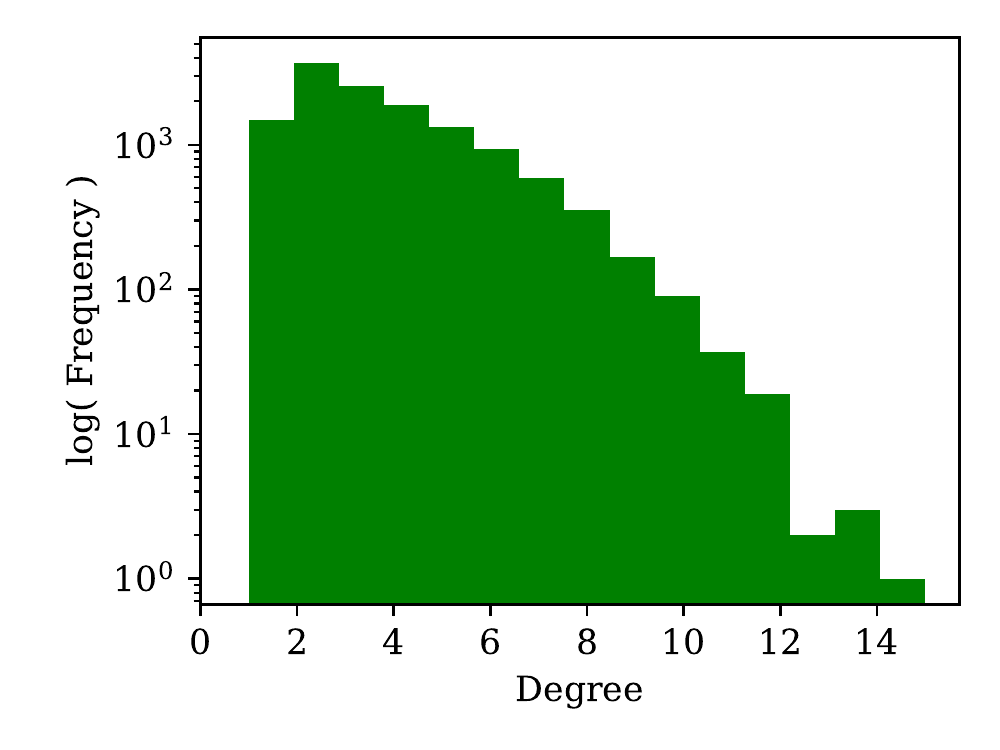}
         \caption{\sgbase dataset degree distribution.}
         \label{fig:sgbase_ddist}
     \end{subfigure}
     \hfill
     \begin{subfigure}[b]{0.45\textwidth}
         \centering
         \includegraphics[width=0.81\textwidth]{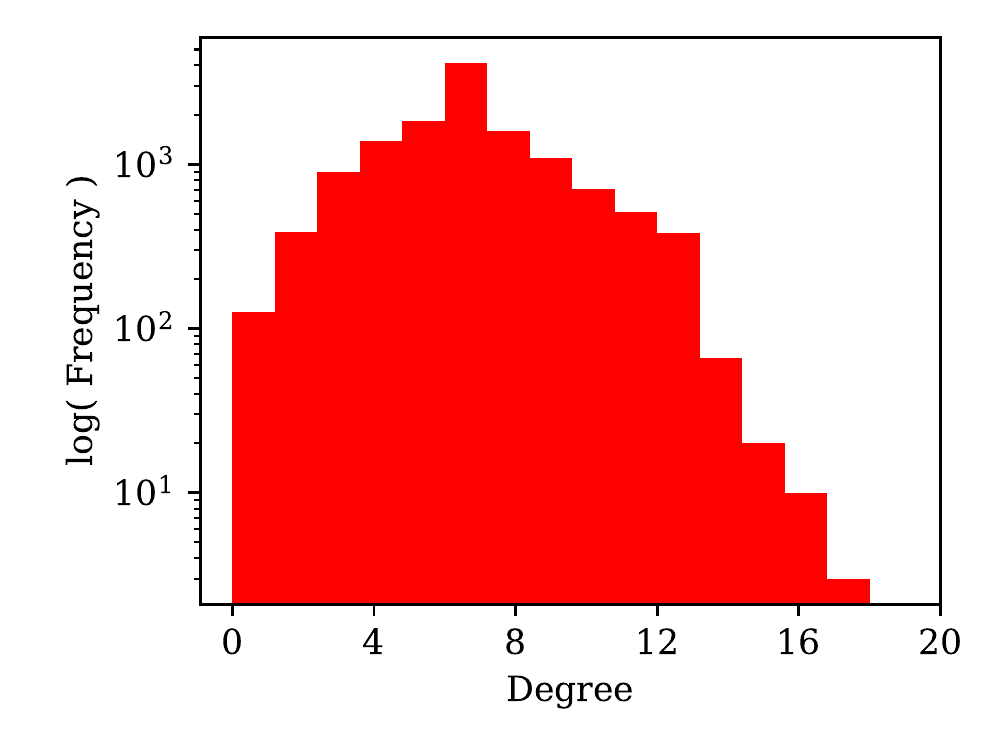}
         \caption{Erd\H{o}s-R\'{e}nyi graph ($p = 5\times10^{-4}$) degree distribution.}
         \label{fig:er_ddist}
     \end{subfigure}
     % \hfill
     \begin{subfigure}[b]{0.45\textwidth}
         \centering
         \includegraphics[width=0.81\textwidth]{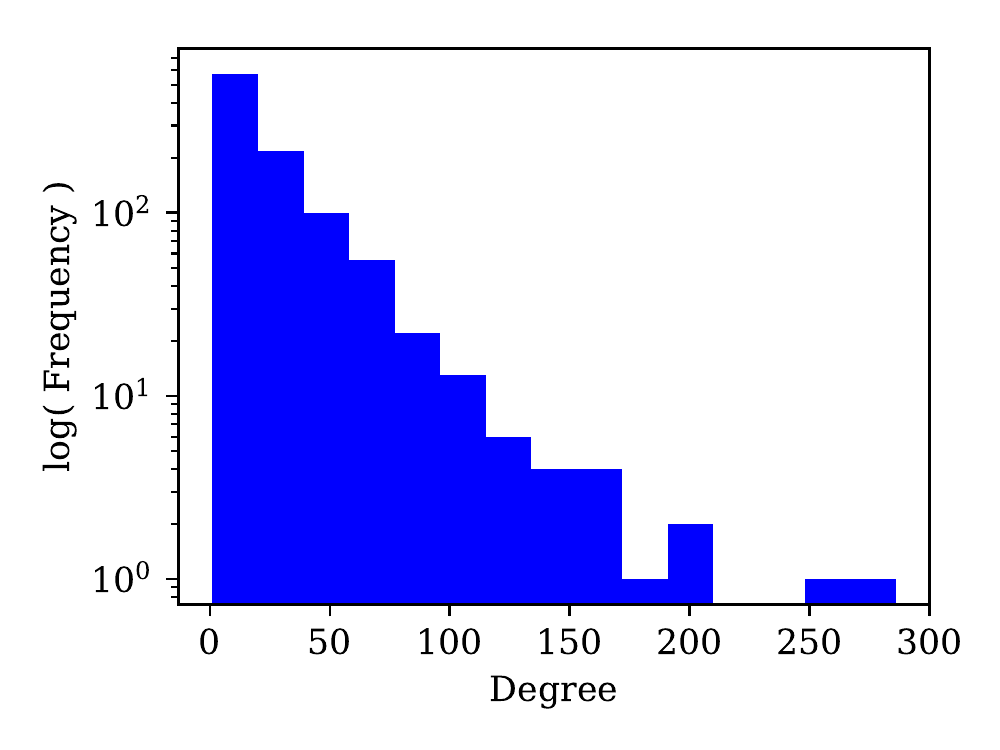}
         \caption{German credit dataset degree distribution.}
         \label{fig:german_ddist}
     \end{subfigure}
     \hfill
     \begin{subfigure}[b]{0.45\textwidth}
         \centering
         \includegraphics[width=0.81\textwidth]{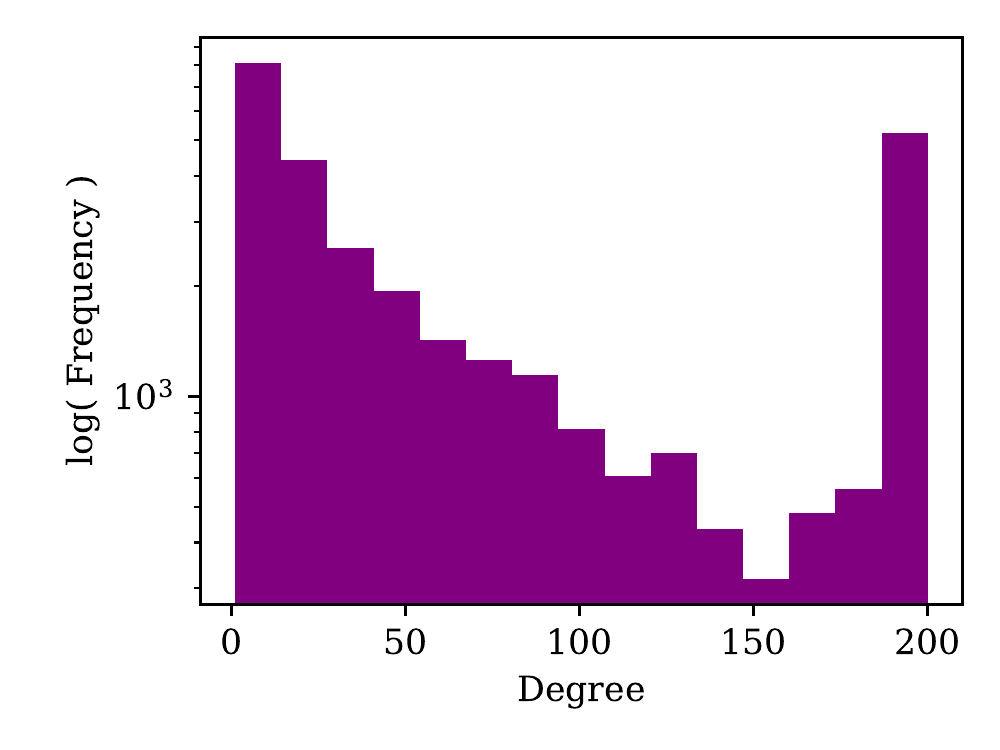}
         \caption{Credit defaulter dataset degree distribution.}
         \label{fig:credit_ddist}
     \end{subfigure}
    \caption{Comparison of degree distribution for a (\subref{fig:sgbase_ddist}) \datagen dataset (\sgbase), (\subref{fig:er_ddist}) a random baseline graph, and two real-world datasets: (\subref{fig:german_ddist}) German Credit and (\subref{fig:credit_ddist}) Credit Defaulter. All plots are shown with a log scale of frequency for the y-axis. \sgbase and both real-world graphs show a power-law degree distribution commonly observed in real-world datasets exhibiting preferential attachment properties. Datasets generated by \datagen are designed to present power-law degree distributions to match real-world dataset topologies, such as those observed in German Credit and Credit Defaulter. The degree distribution of \sgbase is much different than the binomial distribution exhibited in Erd\H{o}s-R\'{e}nyi graph (\subref{fig:er_ddist}), an unstructured random graph model.}
        \label{fig:three graphs}
\end{figure}

\restoregeometry

\clearpage

\captionsetup{margin=0.1in}

\begin{algorithm}
\SetAlgoLined
\footnotesize
\DontPrintSemicolon
\caption{Overview of \datagen Algorithm}
\textbf{Input}: Shape $\mathcal{S}=(\mathcal{V}_{\mathcal{S}}, \mathcal{E}_{\mathcal{S}})$; number of subgraphs $N_s$; probability of connection $p$; subgraph size $n_s$; number of classes $K$; number of features $n_{\text{f}}$; number of informative features $n_{\text{i}}$; class separation factor $s_{\text{f}}$;  number of clusters per class $c_{\text{f}}$; protected feature noise factor $\phi$; homophily coefficient $\eta$; model layers $L$\\

$\mathcal{G} \gets \textsc{ShapeGGenStructure}$($\mathcal{S}$, $N_s$, $p$, $n_s$, $K$) (See Algorithm~\ref{alg:shapegraphstructure}) \\ 
$\bY \gets$ Labels for each node $v \in \mathcal{G}$ by Equation (\ref{eq:shapegraph_f})\\
$\bX, \textbf{M}_F \gets$ \textsc{NodeFeatureVectors}($\mathcal{V}_{\mathcal{G}}$, $\bY$, $K$, $n_{\text{f}}$, $n_{\text{i}}$, $s_{\text{f}}$, $c_{\text{f}}$, $\phi$, $\eta$) \\

$\textbf{M}_N, \textbf{M}_E \gets \textsc{CreateGroundTruthExplanations} (\cG, L)$ \\

\hrulefill \\

\textsc{\textbf{NodeFeatureVectors}}\\
\textbf{Input:} Nodes $\mathcal{V}$; labels of all nodes $\bY$; number of classes/labels $K$; number of features $n_{\text{f}}$; number of informative features $n_{\text{i}}$; separation factor $s_{\text{f}}$; number of clusters per class $c_{\text{f}}$; protected feature noise factor $\phi$; homophily coefficient $\eta$\\
\textbf{Output:} Node features $\bX$, Ground-truth feature explanation $\mathbf{M}_F$\\

Partition $\mathcal{V}$ into $K$ sets, separated by labels,$\{ \mathcal{V}_1, ..., \mathcal{V}_K \}$ \\
Generate hypercube $\mathbf{H}$ in $n_{\text{i}}$ dimensions scaled by size $s_{\text{f}}$\\
Sample $K \times c_{\text{f}}$ vertices $\{ h_1, ... h_K\}$ from $\mathbf{H}$ such that $h_i \in \mathbb{R}^{n_{\text{i}}}$\\
\For{$i\gets1$ \KwTo $K$}{
    Assign features $x \sim \mathcal{N}(h_i, 1)$ to each node $v \in \mathcal{V}_i$ \\
    Noise vectors ${r \sim \mathcal{N}(0, 1)}$, $r \in \mathbb{R}^{n_{\text{f}} - n_{\text{i}}}$ \\
    Set protected feature $q$ equal to label $i$ \\
    For each node, flip $q$ with probability of $\phi$ to another random label\\
    $\bX[v_j] \gets x[v_j] \oplus r[v_j] \oplus q[v_j] ~ \forall \ v_j \in \mathcal{V}_i$ 
}
$\mathbf{M}_F \gets $ Mask over all informative dimensions of features\\
%$\bX$ features for each node in $\mathcal{V}$\\ 
$\bX \gets$ \textsc{OptimizeHomophilyHeterophily}($\mathcal{G}, \bX, \bY, \eta$) in Algorithm~\ref{alg:optimizehomophily}\\
Return $\bX, \textbf{M}_F$

\hrulefill \\
\textsc{\textbf{CreateGroundTruthExplanations}} \\
\textbf{Input}: \datagen $\mathcal{G}=(\mathcal{V}_{\mathcal{G}}, \mathcal{E}_{\mathcal{G}})$; number of layers $L$\\
\textbf{Output:} Ground-truth node explanation $\mathbf{M}_{N}$ and edge explanation $\mathbf{M}_{E}$ \\

Extract all planted motifs in $\mathcal{G}$, $\mathbf{S} = \{ \mathcal{S}_1, \dots, \mathcal{S}_N\}$\\ 
Define $\textsc{Sub}: \mathcal{V}_{\mathcal{G}} \mapsto (\mathcal{V_{\textsc{Sub}}}, \mathcal{E_{\textsc{Sub}}})$ as $L$-hop subgraph around $v_i \in \mathcal{V}_{\mathcal{G}}$ \\
\For{$i\gets1$ \KwTo $|\mathcal{V}_{\mathcal{G}}|$}{
    $\mathcal{V_{\textsc{Sub}}} \gets$ Nodes in $\textsc{Sub}(v_i)$; $\mathcal{E_{\textsc{Sub}}} \gets$ Edges in $\textsc{Sub}(v_i)$ \\
    $\mathbf{S}_{\textsc{Sub}(v_i)} \gets \textsc{Sub}(v_i) \cap \big( \bigcup_{j = 1}^{|\mathbf{S}|} \cS_j \big)$ \\
    
    \For{$j\gets1$ \KwTo $|\textsc{Sub}(v_i)|$}{
        $\mathbf{M}_{N}^{v_i}[j] \gets$ 1 if $v_j \in \mathbf{S}_{\textsc{Sub}(v_i)}$, else 0
    }
    $\mathbf{M}_{E}^{v_i} \gets$ 1 for any edge $(v_a, v_b)$ such that $\mathbf{M}_{N}^{v_i}[a] = \mathbf{M}_{N}^{v_i}[b] = 1$, else 0 
}
Return $\textbf{M}_N, \textbf{M}_E$
\label{alg:shapegraph}
\end{algorithm}

\begin{algorithm}[ht]
\caption{\textsc{ShapeGGenStructure}}
\label{alg:shapegraphstructure}
\SetAlgoLined
\footnotesize
\DontPrintSemicolon
\textbf{Input}: Shape $\mathcal{S}=(\mathcal{V}_{\mathcal{S}}, \mathcal{E}_{\mathcal{S}})$; number of subgraphs $N_s$; probability of connection $p$; subgraph size $n_s$; number of classes $K$\\
\textbf{Output}: \datagen $\mathcal{G}$\\ 
%Make list $\mathbf{S}$ using $N_s$ copies of $\mathcal{S}$ \\
$\mathbf{S} \gets \{\mathcal{S}_1, \mathcal{S}_2, ..., \mathcal{S}_{N_s}\}$\\
\For{$i\gets1$ \KwTo $N_s$}{
    $\mathcal{V}_{\mathcal{S}_i} \gets$ nodes in $\mathcal{S}_i$\\
    $\Delta s_i \gets n_s - |\mathcal{V}_{\mathcal{S}_i}|$ \\
    $n_i \sim \textrm{Poisson}(\lambda = \Delta s_i)$ \\
    \For{$j\gets1$ \KwTo $n_i$}{ 
        $g_i \propto \frac{\textrm{deg}(v)}{\sum_{v_j \in V_{\mathcal{S}_i}} \textrm{deg}(v_j)} \forall v \in V_{\mathcal{S}_i}$ \\
        $v^{*} \gets$ Node sampled with $g_i$ from $V_{\mathcal{S}_i}$ \\ 
        Introduce new node $v'$ to subgraph $\mathcal{S}_i$ with edge $(v', v^*)$
    }
}
$\mathcal{G} \gets \bigcup\limits_{i = 1}^{N_s} \mathcal{S}_i$ \\
\For{$i\gets1$ \KwTo $N_s$}{
    \For{$j\gets i$ \KwTo $N_s$}{
        \eIf{$p$}
        {continue}
        {   
            $\mathcal{E}_{ij} \gets$ All possible edges that could connect $\mathcal{S}_i$ and $\mathcal{S}_j$\\
            \While{$\mathcal{E}_{ij} \neq \emptyset$}{
                $g_{ij} \propto  \frac{\textrm{deg}(v_1) + \textrm{deg}(v_2)}{\sum\limits_{(v_k, v_m) \in \mathcal{E}_{ij}}\textrm{deg}(v_k) + \textrm{deg}(v_m)}\ \forall \ (v_1, v_2) \in \mathcal{E}_{ij}$ \\
                $e^{*} \gets $ Edge sampled with $g_{ij}$ from $\mathcal{E}_{ij}$ \\x
                Remove $e^{*}$ from $\mathcal{E}_{ij}$ \\
                Add edge $e^{*}$ to $\mathcal{G}$  \\
                \eIf{ $\exists~v \in \mathcal{V}_{\mathcal{G}}$ \text{such that} $f(v) > K - 1$ (Eqn. \ref{eq:shapegraph_f})}{
                   Remove $e^{*}$ from $\mathcal{G}$
                }
                {
                    Break
                }
            }
        }
    }
}
$\mathcal{G}^{*} \gets$ Largest connected component of $\mathcal{G}$\\
Return $\mathcal{G}^*$
\end{algorithm}
% \hrulefill \\

\begin{algorithm}[ht]
\footnotesize
\textbf{Input:} Graph $\mathcal{G}$, node features $\bX$, node labels $\bY$, homophily coefficient $\eta$ \\
\textbf{Output:} Optimized node features $\bX_{\eta}$ \\
$\mathbf{C} \gets$ Sample connected set of nodes with same labels in $\bY$\\
$\mathbf{D}_{\text{dis}} \gets$ Sampled set of \textit{disconnected} pairs of nodes with different labels\\
$\mathbf{D}_{\text{con}} \gets$ Sampled set of \textit{connected} pairs of nodes with different labels \\

$ L_{\mathbf{C}}(\bX) = -\eta \frac{1}{|\mathbf{C}|}\sum_{(v_i, v_j) \in \mathbf{C}} D(\bX[v_i], \bX[v_j])$ \\

$L_{\mathbf{dis}}(\bX) = \eta \frac{1}{|\mathbf{D_{dis}}|}\sum_{(v_i, v_j) \in \mathbf{D_{dis}}} D(\bX[v_i], \bX[v_j])$ \\

$ L_{\mathbf{con}}(\bX) = \eta \frac{1}{|\mathbf{D_{con}}|}\sum_{(v_i, v_j) \in \mathbf{D_{con}}} D(\bX[v_i], \bX[v_j])$ \\

$L(\bX) = L_{\mathbf{C}}(\bX) + L_{\mathbf{dis}}(\bX) + L_{\mathbf{con}}(\bX)$ \\
Return $\argmin\limits_{\bX} L(\bX)$
\caption{\textsc{{OptimizeHomophilyHeterophily}}}
\label{alg:optimizehomophily}
\end{algorithm}

\clearpage

\begin{table}[t]
\centering
% \renewcommand{\arraystretch}{0.9}
% \setlength{\tabcolsep}{1.2pt}
% \vspace{-1mm}
{\begin{tabular}{l|ccccc}
%  & \multicolumn{5}{c}{Performance Metrics} \\
{Method} & {GEA ($\uparrow$)} & {GEF ($\downarrow$)} & {GES ($\downarrow$)} & {GECF ($\downarrow$)} & {GEGF ($\downarrow$)} \\
\toprule
\begin{tabular}[l]{@{}l@{}}{Random}\\{{Grad}}\\{{GradCAM}}\\{GuidedBP}\\{IG}\\{{GNNExplainer}}\\{{PGMExplainer}}\\{PGExplainer}\\{SubgraphX}\end{tabular} & % {CAM}\\
% GEA:
\begin{tabular}[c]{@{}c@{}}{0.148}\std{0.002}\\{0.193}\std{0.002}\\{0.222}\std{0.002}\\{0.194}\std{0.001}\\{0.142}\std{0.002}\\{0.102}\std{0.003}\\{0.133}\std{0.002}\\{0.194}\std{0.002}\\\textbf{0.324}\std{0.004}\end{tabular} & % \\{0.189}\std{0.002}
% GEF:
\begin{tabular}[c]{@{}c@{}}{0.579}\std{0.007}\\{0.392}\std{0.006}\\{0.452}\std{0.006}\\{0.557}\std{0.007}\\{0.545}\std{0.007}\\{0.534}\std{0.007}\\{0.541}\std{0.007}\\{0.557}\std{0.007}\\\textbf{0.254}\std{0.006}\end{tabular} & % \\{0.469}\std{0.006}
%GES:
\begin{tabular}[c]{@{}c@{}}{0.920}\std{0.002}\\{0.806}\std{0.004}\\{0.263}\std{0.004}\\{0.432}\std{0.004}\\{0.727}\std{0.005}\\{0.431}\std{0.008}\\{0.984}\std{0.001}\\\textbf{0.217}\std{0.004}\\{0.745}\std{0.005}\end{tabular} & % {0.279}\std{0.004}\\
% \begin{tabular}[c]{@{}c@{}}{}\\{}\\{}\\{}\\{}\\{}\\{}\\{}\\{}\end{tabular} &
% GECF:
\begin{tabular}[c]{@{}c@{}}{0.763}\std{0.003}\\{0.159}\std{0.004}\\{0.010}\std{0.001}\\{0.067}\std{0.002}\\{0.110}\std{0.003}\\{0.233}\std{0.006}\\{0.791}\std{0.003}\\\textbf{0.009}\std{0.000}\\{0.241}\std{0.006}\end{tabular} &  % {0.033}\std{0.003}\\

% GEGF:
\begin{tabular}[c]{@{}c@{}}{0.023}\std{0.002}\\{0.039}\std{0.003}\\\textbf{0.020}\std{0.002}\\{0.021}\std{0.002}\\{0.021}\std{0.002}\\{0.027}\std{0.002}\\{0.096}\std{0.004}\\{0.029}\std{0.002}\\{0.035}\std{0.003}\end{tabular} % {0.023}\std{0.002}\\
 \\
\bottomrule
\end{tabular}}
\caption{
    Evaluation of GNN explainers on \sgbase graph dataset based on node explanation masks $\bM^{p}_{N}$. Arrows ($\uparrow$/$\downarrow$) indicate the direction of better performance. SubgraphX far outperforms other methods in accuracy and faithfulness while PGExplainer is best for stability and counterfactual fairness. In general, gradient methods produce the most fair explanations across both counterfactual and group fairness metrics. See Table~\ref{tab:benchmark_edge}-\ref{tab:benchmark_feat} for results on edge and feature explanation masks.
}
\label{tab:benchmark_node}
\end{table}

\begin{table}[!ht]
\centering
% \small
% \renewcommand{\arraystretch}{0.9}
% \setlength{\tabcolsep}{1.3pt}
% \vspace{-1mm}
{\begin{tabular}{l|cccccc}
%  & \multicolumn{5}{c}{Performance Metrics} \\
{Method} & {GEA ($\uparrow$)} & {GEF ($\downarrow$)} & {GES ($\downarrow$)} & {GECF ($\downarrow$)} & {GEGF ($\downarrow$)} \\
\toprule
\begin{tabular}[l]{@{}l@{}}{Random}\\{{Grad}}\\{{GradCAM}}\\{GuidedBP}\\{IG}\\{{GNNExplainer}}\\{{PGMExplainer}}\\{PGExplainer}\\{SubgraphX}\end{tabular} & 
% GEA (0.1*max):
% \begin{tabular}[c]{@{}c@{}}{0.074}\std{0.002}\\{0.190}\std{0.002}\\{0.183}\std{0.001}\\{0.185}\std{0.001}\\{0.136}\std{0.002}\\{0.101}\std{0.003}\\{0.130}\std{0.002}\\{}\std{}\\{}\std{}\end{tabular} & % \\{0.189}\std{0.002}
% GEA (top-k):
\begin{tabular}[c]{@{}c@{}}{0.075}\std{0.002}\\{0.194}\std{0.002}\\{0.188}\std{0.001}\\{0.190}\std{0.001}\\{0.140}\std{0.002}\\{0.103}\std{0.003}\\{0.133}\std{0.002}\\{0.165}\std{0.002}\\\textbf{0.383}\std{0.004}\end{tabular} &
% GEF:
\begin{tabular}[c]{@{}c@{}}{0.638}\std{0.007}\\{0.498}\std{0.007}\\{0.620}\std{0.006}\\{0.653}\std{0.007}\\{0.672}\std{0.007}\\{0.632}\std{0.007}\\{0.622}\std{0.007}\\{0.635}\std{0.007}\\\textbf{0.344}\std{0.006}\end{tabular} & % \\{0.189}\std{0.002}
% GES:
\begin{tabular}[c]{@{}c@{}}{1.55}\std{0.004}\\{0.745}\std{0.005}\\{0.295}\std{0.005}\\{0.430}\std{0.004}\\{0.639}\std{0.004}\\{0.431}\std{0.008}\\{0.974}\std{0.001}\\\textbf{0.224}\std{0.004}\\{0.585}\std{0.004}\end{tabular} & 
% GECF:
\begin{tabular}[c]{@{}c@{}}{1.01}\std{0.010}\\{0.157}\std{0.004}\\{0.029}\std{0.003}\\{0.074}\std{0.003}\\{0.114}\std{0.004}\\{0.249}\std{0.007}\\{0.798}\std{0.003}\\\textbf{0.005}\std{0.000}\\{0.225}\std{0.004}\end{tabular} & 
% GEGF:
\begin{tabular}[c]{@{}c@{}}{0.027}\std{0.002}\\{0.068}\std{0.003}\\{0.027}\std{0.002}\\{0.020}\std{0.002}\\\textbf{0.011}\std{0.001}\\{0.028}\std{0.002}\\{0.083}\std{0.003}\\{0.030}\std{0.002}\\{0.114}\std{0.004}\end{tabular}
 \\
\bottomrule
\end{tabular}}
\caption{
    Evaluation of GNN explainers on \sgbase graph dataset based on node explanation masks $\bM^{p}_{N}$. Base GNN is a GCN \cite{kipf17:semi} as opposed to Table \ref{tab:benchmark_node} which is based on explaining a GIN model \cite{xu2018powerful}. Overall, explainer performance is very similar to that of the GIN with SubgraphX performing the best on faithfulness and accuracy metrics while gradient-based methods and PGExplainer typically perform best for fairness and stability.
}
\label{tab:gcn_benchmark_node}
\end{table}

\begin{table}[!ht]
\centering%\small
% \renewcommand{\arraystretch}{1.1}
% \setlength{\tabcolsep}{1.2pt}
% \vspace{-1mm}
{\begin{tabular}{l|ccccc}
%  & \multicolumn{5}{c}{Performance Metrics} \\
{Method} & {GEA ($\uparrow$)} & {GEF ($\downarrow$)} & {GES ($\downarrow$)} & {GECF ($\downarrow$)} & {GEGF ($\downarrow$)} \\
\toprule
\begin{tabular}[l]{@{}l@{}}{Random}\\{{GNNExplainer}}\\{PGExplainer}\\{SubgraphX}\end{tabular} & % {CAM}\\
%GEA
\begin{tabular}[c]{@{}c@{}}{0.135}\std{0.001}\\{0.152}\std{0.002}\\{0.117}\std{0.002}\\\textbf{0.271}\std{0.003}\end{tabular} & % \\{0.189}\std{0.002}
%GEF
\begin{tabular}[c]{@{}c@{}}{0.419}\std{0.007}\\{0.302}\std{0.006}\\\textbf{0.171}\std{0.004}\\{0.548}\std{0.007}\end{tabular}  & % \\{0.469}\std{0.006}
%GES
\begin{tabular}[c]{@{}c@{}}{1.167}\std{0.001}\\{0.995}\std{0.001}\\{1.000}\std{0.000}\\\textbf{0.815}\std{0.005}\end{tabular} & % {0.279}\std{0.004}\\
%GCF
\begin{tabular}[c]{@{}c@{}}{0.997}\std{0.002}\\{0.957}\std{0.002}\\{1.000}\std{0.000}\\\textbf{0.290}\std{0.007}\end{tabular} &  % {0.033}\std{0.003}\\
%GGF
\begin{tabular}[c]{@{}c@{}}{0.064}\std{0.003}\\{0.047}\std{0.003}\\{0.037}\std{0.003}\\\textbf{0.030}\std{0.002}\end{tabular}
 \\
\bottomrule
\end{tabular}}
\caption{Evaluation of GNN explainers on \sgbase graph dataset based on edge explanation masks $\bM^{p}_{E}$. Arrows ($\uparrow$/$\downarrow$) indicate the direction of better performance. SubgraphX method, on average, produces the most reliable edge explanations when evaluated across all five performance metrics. Note that of the explainers tested in this study, only the above four methods produce edge explanations.}
\label{tab:benchmark_edge}
\vskip -0.1in
\end{table}

\begin{table}[!ht]
\centering%\small
% \renewcommand{\arraystretch}{1.1}
% \setlength{\tabcolsep}{1.2pt}
% \vspace{-1mm}
{\begin{tabular}{l|ccccc}
%  & \multicolumn{5}{c}{Performance Metrics} \\
{Method} & {GEA ($\uparrow$)} & {GEF ($\downarrow$)} & {GES ($\downarrow$)} & {GECF ($\downarrow$)} & {GEGF ($\downarrow$)} \\
\toprule
\begin{tabular}[l]{@{}l@{}}{Random}\\{{Grad}}\\{GuidedBP}\\{IG}\\{{GNNExplainer}}\end{tabular} & % {CAM}\\
% GEA
\begin{tabular}[c]{@{}c@{}}{0.281}\std{0.003}\\\textbf{0.306}\std{0.002}\\{0.240}\std{0.003}\\{0.278}\std{0.003}\\{0.281}\std{0.003}\end{tabular}  & % \\{0.189}\std{0.002}
%GEF
\begin{tabular}[c]{@{}c@{}}{0.016}\std{0.001}\\\textbf{0.015}\std{0.001}\\{0.016}\std{0.001}\\{0.016}\std{0.001}\\{0.017}\std{0.001}\end{tabular}  & % \\{0.469}\std{0.006}
%GES
\begin{tabular}[c]{@{}c@{}}{0.997}\std{0.001}\\{0.925}\std{0.003}\\\textbf{0.899}\std{0.004}\\{0.917}\std{0.004}\\{0.999}\std{0.001}\end{tabular}  & % {0.279}\std{0.004}\\
% \begin{tabular}[c]{@{}c@{}}{}\\{}\\{}\\{}\\{}\\{}\\{}\\{}\\{}\end{tabular} &
%GCF
\begin{tabular}[c]{@{}c@{}}{0.810}\std{0.005}\\{0.259}\std{0.006}\\{0.275}\std{0.006}\\\textbf{0.119}\std{0.004}\\{0.826}\std{0.005}\end{tabular}  &  % {0.033}\std{0.003}\\
%GGF
\begin{tabular}[c]{@{}c@{}}{0.023}\std{0.002}\\{0.027}\std{0.003}\\{0.025}\std{0.002}\\\textbf{0.022}\std{0.002}\\{0.023}\std{0.003}\end{tabular}  % {0.023}\std{0.002}\\
 \\
\bottomrule
\end{tabular}}
\caption{
Evaluation of GNN explainers on \sgbase graph dataset based on node feature explanation masks $\bM^{p}_{F}$. Arrows ($\uparrow$/$\downarrow$) indicate the direction of better performance. All GNN explainers produce highly faithful node feature explanations. However, the stability of these methods on node features is more similar to random explanations than is observed for node explanations in Table \ref{tab:benchmark_node} and Table \ref{tab:gcn_benchmark_node}. Note that of the explainers tested in this study, only the above five methods produce node feature explanations.}
\label{tab:benchmark_feat}
\vskip -0.1in
\end{table}

\begin{table}[!ht]
    \centering
    \begin{tabular}{ll|cc}
        Dataset & Method & GEA~($\uparrow$) & GEF~($\downarrow$)\\
        \toprule
        \multirow{9}{*}{\textsc{Mutag}} & Random & {0.044}\std{0.007} & {0.590}\std{0.031} \\ 
        & Grad & {0.022}\std{0.006} & {0.598}\std{0.030}\\
        % & CAM & {0.074}\std{0.014} & {0.654}\std{0.030}\\
        & GradCAM & \textbf{0.085}\std{0.012} &  {0.672}\std{0.029} \\
        & GuidedBP & {0.036}\std{0.007} & {0.649}\std{0.030} \\
        & Integrated Grad (IG) & {0.049}\std{0.010} & \textbf{0.443}\std{0.031} \\
        & GNNExplainer & {0.031}\std{0.005} & {0.618}\std{0.030} \\
        & PGMExplainer & {0.042}\std{0.007} & {0.503}\std{0.031} \\
        & PGExplainer & {0.046}\std{0.007} & {0.504}\std{0.031} \\
        & SubgraphX & {0.039}\std{0.007} & {0.611}\std{0.030} \\
        \midrule % ------------------------
        \multirow{9}{*}{\textsc{Benzene}} & Random & {0.108}\std{0.003} & {0.513}\std{0.012} \\ 
        & Grad & {0.122}\std{0.007} & {0.262}\std{0.011} \\
        % & CAM & 0.250\std{0.007} & 0.534\std{0.012}\\
        & GradCAM & {0.291}\std{0.007} & {0.551}\std{0.012} \\
        & GuidedBP & {0.205}\std{0.007} & {0.438}\std{0.012} \\
        & Integrated Grad (IG) & {0.044}\std{0.003} & \textbf{0.182}\std{0.010} \\
        & GNNExplainer & {0.129}\std{0.005} & {0.444}\std{0.012} \\
        & PGMExplainer & {0.154}\std{0.006} & {0.433}\std{0.012} \\
        & PGExplainer & {0.169}\std{0.007} & {0.375}\std{0.012} \\
        & SubgraphX & \textbf{0.371}\std{0.009} & {0.513}\std{0.012} \\
        \midrule
        \multirow{9}{*}{\textsc{Fl-Carbonyl}} & Random & {0.087}\std{0.007} & 0.440\std{0.26}\\ 
        & Grad & \textbf{0.132}\std{0.010} & 0.210\std{0.021}\\
        & GradCAM & {0.005}\std{0.007} & 0.500\std{0.026}\\
        & GuidedBP & {0.089}\std{0.010} & 0.315\std{0.024}\\
        & Integrated Grad~(IG) & {0.091}\std{0.007} & \textbf{0.174}\std{0.019}\\
        & GNNExplainer & {0.094}\std{0.009} & {0.423}\std{0.026} \\
        & PGMExplainer & {0.078}\std{0.008} & 0.426\std{0.026} \\
        & PGExplainer & {0.079}\std{0.009} & {0.372}\std{0.025} \\
        & SubgraphX & {0.008}\std{0.002} & 0.466\std{0.026} \\
        \bottomrule
    \end{tabular}
    \caption{Evaluation of GNN explainers for real-world molecular datasets with ground-truth explanations. Arrows ($\uparrow$/$\downarrow$) indicate the direction of better performance. Integrated Gradient explanations obtain the lowest unfaithfulness score across all three datasets. Note that stability and fairness do not apply to these datasets because generating plausible perturbations for molecules is non-trivial, and they do not contain protected features.
    }
    \vskip -0.2in
    \label{tab:benchmark_graph}
\end{table}

\hide{
    \begin{table}[]
\centering
\begin{tabular}{@{}lrr@{}}
\toprule
Node Type          & Count   & Percent (\%) \\ \midrule
Biological process & 28,642  & 22.1         \\
Protein            & 27,671  & 21.4         \\
Disease            & 17,080  & 13.2         \\
Phenotype          & 15,311  & 11.8         \\
Anatomy            & 14,035  & 10.8         \\
Molecular function & 11,169  & 8.6          \\
Drug               & 7,957   & 6.2          \\
Cellular component & 4,176   & 3.2          \\
Pathway            & 2,516   & 1.9          \\
Exposure           & 818     & 0.6          \\
\midrule
Total              & 129,375 & 100.0        \\ \bottomrule
\end{tabular}
\caption{Statistics on nodes in \method.}
\label{tab:kg_nodes}
\end{table}

    \begin{table}[]
\centering
\begin{tabular}{@{}lrr@{}}
\toprule
Relation type                                & Count     & Percent (\%) \\
\midrule
Anatomy - Protein (present)             & 3,036,406 & 37.5         \\
Drug - Drug                             & 2,672,628 & 33.0         \\
Protein - Protein                       & 642,150   & 7.9          \\
Disease - Phenotype (positive)          & 300,634   & 3.7          \\
Biological process - Protein            & 289,610   & 3.6          \\
Cellular component - Protein            & 166,804   & 2.1          \\
Disease - Protein                       & 160,822   & 2.0          \\
Molecular function - Protein            & 139,060   & 1.7          \\
Drug - Phenotype                        & 129,568   & 1.6          \\
Biological process - Biological process & 105,772   & 1.3          \\
Pathway - Protein                       & 85,292    & 1.1          \\
Disease - Disease                       & 64,388    & 0.8          \\
Drug - Disease (contraindication)       & 61,350    & 0.8          \\
Drug - Protein                          & 51,306    & 0.6          \\
Anatomy - Protein (absent)              & 39,774    & 0.5          \\
Phenotype - Phenotype                   & 37,472    & 0.5          \\
Anatomy - Anatomy                       & 28,064    & 0.3          \\
Molecular function - Molecular function & 27,148    & 0.3          \\
Drug - Disease (indication)             & 18,776    & 0.2          \\
Cellular component - Cellular component & 9,690     & 0.1          \\
Phenotype - Protein                     & 6,660     & 0.1          \\
Drug - Disease (off-label use)          & 5,136     & 0.1          \\
Pathway - Pathway                       & 5,070     & 0.1          \\
Exposure - Disease                      & 4,608     & 0.1          \\
Exposure - Exposure                     & 4,140     & 0.1          \\
Exposure - Biological process           & 3,250     & \textless \ 0.1          \\
Exposure - Protein                      & 2,424     & \textless \ 0.1          \\
Disease - Phenotype (negative)          & 2,386     & \textless \ 0.1          \\
Exposure - Molecular function           & 90        & \textless \ 0.1          \\
Exposure - Cellular component           & 20        & \textless \ 0.1          \\
\midrule
Total                                   & 8,100,498 & 100.0  \\
\bottomrule
\end{tabular}
\caption{Statistics on edges in \method.}
\label{tab:kg_edges}
\end{table}
}

% \usepackage{booktabs}

\begin{table}[!ht]
    \centering\small
    \setlength{\tabcolsep}{3.5pt}
    \begin{tabular}{lcccc}
    \toprule
    Dataset & \sgbase & \sghetero & \sgsmallex & \sgunfair\\ \midrule 
           Nodes & 13150 & 13150 & 15505 & 13150 \\
           Edges & 46472 & 46472 & 51782 & 46472  \\
           Node features & 11 & 11 & 11 & 11\\
           Average node degree & 3.53\std{0.02} & 3.53\std{0.02} & 3.34\std{0.02} & 3.53\std{0.02}\\
           Class 0 Nodes & 4382 & 4382 & 7777 & 4382\\
           Class 1 Nodes & 8768 & 8768 & 7728 & 8768\\
           \midrule
           Shape of the planted motif ($\mathcal{S}$) & `house' & `house' & `triangle' & `house'\\
           Number of initial subgraphs ($N_{s}$) & 1200 & 1200 & 1300 & 1200\\
           Probability of subgraph connection ($p$) & 0.006 & 0.006 & 0.006 & 0.006\\
           Subgraph size ($n_s$) & 11 & 11 & 12 & 11\\
           Number of classes ($K$) & 2 & 2 & 2 & 2\\
           Number of node features ($n_{{f}}$) & 11 & 11 & 11 & 11 \\
           Number of informative features ($n_{{i}}$) & 4 & 4 & 4 & 4 \\
           Class separation factor ($s_{{f}}$) & 0.6 & 0.6 & 0.5 & 0.6 \\
           Number of clusters per class ($c_{{f}}$) & 2 & 2 & 2 & 2 \\
           Protected feature noise factor ($\phi$) & 0.5 & 0.5 & 0.5 & 0.75 \\
           Homophily coefficient ($\eta$) & 1 & -1 & 1 & 1\\
           Number of GNN layers ($L$) & 3 & 3 & 3 & 3\\
    \bottomrule
    \end{tabular}
    \caption{Statistics of graphs generated using \datagen data generator for evaluating different properties of GNN explanations. 
    } 
    \label{tab:data-graph}
\end{table}

\begin{table}[!ht]
    \centering\small
    \begin{tabular}{lcccc}
    \toprule
    Dataset & MUTAG & Benzene & Fluoride-Carbonyl & Alkane-Carbonyl\\ \midrule 
           Graphs & 1,768 & 12,000 & 8,671 & 4,326 \\
           Average nodes & 29.15\std{0.35} & 20.58\std{0.04} & 21.36\std{0.04} & 21.13\std{0.05}\\
           Average edges & 60.83\std{0.75} & 43.65\std{0.08} & 45.37\std{0.09} & 44.95\std{0.12}\\
           Node features & 14 & 14 & 14 & 14 \\
           \midrule
           GT Explanation & \makecell{NH$_{2}$, NO$_{2}$ \\chemical group} & Benzene Ring &  \makecell{F$^{-}$ and C=O \\chemical group} & \makecell{Alkane and C=O\\ chemical group}\\
    \bottomrule
    \end{tabular}
    \caption{
        Statistics of real-world graph classification datasets in \name with ground-truth (GT) explanations.  
        } 
    \label{tab:data-real-graph}
\end{table}

\begin{table}[!ht]
    \centering\small
    \begin{tabular}{lccc}
    \toprule
    Dataset & German credit graph & Recidivism graph & Credit defaulter graph \\ \midrule 
        Nodes & 1,000 & 18,876 & 30,000 \\
        Edges & 22,242 & 321,308 & 1,436,858 \\
        Node features & 27 & 18 & 13 \\
        Average node degree & 44.48\std{26.51} & 34.04\std{46.65} & 95.79\std{85.88} \\
        \bottomrule
    \end{tabular}
    \caption{
        Statistics of real-world node classification datasets in \name without ground-truth (GT) explanations.
        } 
    \label{tab:data-node}
\end{table}

\hide{
% \usepackage{booktabs}
% \usepackage{multirow}
\begin{table}[!ht]
\centering
\begin{tabular}{@{}llrrrr@{}}
\toprule
\multirow{2}{*}{Source}      & \multirow{2}{*}{Type of feature} & \multicolumn{2}{c}{Unprocessed KG} & \multicolumn{2}{c}{Processed KG} \\ \cmidrule(lr){3-4} \cmidrule(lr){5-6}
                             &                          & Count       & Unique       & Count        & Unique        \\
\midrule
Combined                          & Combined                      & 40,068       & 18,152        & 39,800        & 14,252         \\
\cmidrule{1-6}

MONDO Disease Ontology\cite{shefchek_monarch_2020}       & Definition               & 15,238       & 15,238        & 15,238        & 12,001         \\
\cmidrule{1-6}
UMLS\cite{bodenreider_unified_2004}                        & Description              & 28,468       & 8,689         & 25,374        & 6,964          \\
\cmidrule{1-6}

\multirow{5}{*}{Orphanet\cite{PMID:18389888}}    
& Definition               & 6,564        & 6,548         & 6,562         & 5,645          \\
& Prevalence               & 3,989        & 3,989         & 3,500         & 3,430          \\
& Epidemiology             & 2,350        & 2,348         & 2,335         & 2,026          \\
& Clinical description     & 2,294        & 2,292         & 2,293         & 1,972          \\
& Management and treatment & 1,732        & 1,731         & 1,722         & 1,553          \\
\cmidrule{1-6}

\multirow{6}{*}{Mayo Clinic\cite{mayo_foundation_for_medical_education_and_research_mfmer_mayo_2020}} & Symptoms                 & 6,642        & 5,789         & 5,140         & 4,470          \\
                             & Causes                   & 6,629        & 5,776         & 5,128         & 4,459          \\
                             & Risk factors             & 6,284        & 5,501         & 4,898         & 4,299          \\
                             & Complications            & 5,011        & 4,455         & 3,792         & 3,396          \\
                             & Prevention               & 2,529        & 2,273         & 1,907         & 1,776          \\
                             & When to see a doctor     & 5,862        & 5,234         & 4,531         & 4,058   \\
\bottomrule
\end{tabular}
\caption{Statistics on disease features in the knowledge graph. Unprocessed KG refers to the initial knowledge graph assembled from datasets. Processed KG refers to the fully processed \method and includes disease groupings. The count column refers to the number of features, including duplicates, and the unique column refers to the number of unique features.}
\label{tab:kg_feat_diseases}
\end{table}
}

\clearpage 

% \section*{References}
% {
\spacing{0.85}
\bibliographystyle{naturemag}
\bibliography{refs}
% }

\end{document}